\documentclass[runningheads,dvipsnames]{llncs}

\usepackage{graphicx}
\usepackage[misc,geometry]{ifsym}
\usepackage{amsmath,amssymb}
\usepackage{bbold}
\usepackage{mathrsfs}
\usepackage{multirow}
\usepackage{array}
\usepackage{makecell}
\usepackage{tikz}
\usepackage{lipsum}
\usetikzlibrary{arrows.meta, spy, calc, arrows, patterns, patterns.meta}
\usepackage{adjustbox}

\newcommand{\norm}[1]{\left\lVert#1\right\rVert}

\newcolumntype{P}[1]{>{\centering\arraybackslash}p{#1}}

\begin{document}

\title{Multi-Task, Multi-Domain Deep Segmentation with Shared Representations and Contrastive Regularization for Sparse Pediatric Datasets}

\titlerunning{Multi-Task, Multi-Domain Deep Segmentation}

\author{Arnaud Boutillon\inst{1,2}\and
Pierre-Henri Conze\inst{1,2} \and
Christelle Pons\inst{2,3,4} \and\\
Valérie Burdin\inst{1,2} \and
Bhushan Borotikar\inst{2,3,5}}

%index{Boutillon, Arnaud}
%index{Conze, Pierre-Henri}
%index{Pons, Christelle}
%index{Burdin, Valérie}
%index{Borotikar, Bhushan}

\authorrunning{A. Boutillon et al.}

\institute{IMT Atlantique, Brest, France \\ \email{arnaud.boutillon@imt-atlantique.fr} \and
LaTIM UMR 1101, Inserm, Brest, France \and
Centre Hospitalier R\'egional et Universitaire (CHRU) Brest, Brest, France \and
Fondation ILDYS, Brest, France \and
SCMIA, Symbiosis International University, Pune, India}

\maketitle              

\begin{abstract}

Automatic segmentation of magnetic resonance (MR) images is crucial for morphological evaluation of the pediatric musculoskeletal system in clinical practice. However, the accuracy and generalization performance of individual segmentation models are limited due to the restricted amount of annotated pediatric data. Hence, we propose to train a segmentation model on multiple datasets, arising from different parts of the anatomy, in a multi-task and multi-domain learning framework. This approach allows to overcome the inherent scarcity of pediatric data while benefiting from a more robust shared representation. The proposed segmentation network comprises shared convolutional filters, domain-specific batch normalization parameters that compute the respective dataset statistics and a domain-specific segmentation layer. Furthermore, a supervised contrastive regularization is integrated to further improve generalization capabilities, by promoting intra-domain similarity and impose inter-domain margins in embedded space. We evaluate our contributions on two pediatric imaging datasets of the ankle and shoulder joints for bone segmentation. Results demonstrate that the proposed model outperforms state-of-the-art approaches.

% Max 3 keywords

\end{abstract}

\section{Introduction}
\label{sec:introduction}

Segmentation of the pediatric musculoskeletal system serves as an essential pre-processing step to guide clinical decisions, as the generated 3D models of muscles and bones help clinicians evaluate pathology progression and optimally plan therapeutic interventions \cite{balassy_role_2008,hirschmann_artificial_2019,meyer_musculoskeletal_2008}. As manual segmentation is the current standard for delineating pediatric magnetic resonance (MR) images, the implementation of automatic and robust segmentation techniques could reduce analysis time and enhance the reliability of morphological evaluation \cite{balassy_role_2008,hirschmann_artificial_2019,meyer_musculoskeletal_2008}. Recently, convolutional neural networks have demonstrated promising results for automatic segmentation tasks \cite{litjens_survey_2017,lundervold_overview_2019}. However, while the development of supervised deep learning models typically requires large amount of annotated data \cite{litjens_survey_2017,lundervold_overview_2019}, the conception of pediatric imaging datasets is a slow and onerous process \cite{hirschmann_artificial_2019}. Hence, the inherent scarcity of pediatric imaging resources can induce limited generalization capabilities in deep learning models \cite{litjens_survey_2017,lundervold_overview_2019}.

To address the similar problem of limited annotated data for semantic scene labeling in natural images, Fourure et al. proposed to train a single network over the union of multiple datasets, in order to leverage a greater amount of training data \cite{fourure_semantic_2016}. Their model is optimized in a multi-task and multi-domain framework, in which each dataset is defined by its own task (segmentation label-set) and domain (data distribution). Hence, this approach is more generic than traditional domain adaptation techniques which usually focus on domains containing the same set of objects \cite{fourure_semantic_2016}. Following this, Rebuffi et al. \cite{rebuffi_efficient_2018,rebuffi_learning_2017} proposed to employ a model with agnostic filters, as visual primitives may be shared across tasks and domains, and dataset-specific layers which allow task and domain specialization. These approaches, based on shared representations, have been reported to perform at par or superior to traditional independent models \cite{dou_unpaired_2020,karani_lifelong_2018,liu_ms-net_2020,valindria_multi_modal_2018}.

Even though these models integrate domain-specific information, this prior knowledge could be further exploited to improve the generalizability of the shared representation. For instance, Zhu et al. imposed a Gaussian mixture distribution on the shared representation of their image translation model \cite{zhu_cross-domain_2020}, however, such a hypothesis may be too restrictive and lead to a decrease in performance. In representation learning, a good representation can be characterized by the presence of natural clusters corresponding to the classes of the problem \cite{bengio_representation_2013}. Hence, a number of self-supervised representation learning approaches focus on pulling together data-points from the same class and pushing apart negative samples in embedded space using a contrastive metric \cite{chen_simple_2020,hadsell_dimensionality_2006}. Khosla et al. extended this idea to fully-supervised classification setting by leveraging the label information and considering many positives simultaneously \cite{khosla_supervised_2020}. Thus, the contrastive regularization maximizes the performance of the classifier by imposing intra-class cohesion and inter-class separation in latent space.

In this study, we propose to develop and optimize a single segmentation network over the union of pediatric imaging datasets acquired on separate anatomical regions. The contributions of this study are threefold: 1) We formalize a segmentation model which incorporates shared representations, domain-specific batch normalization \cite{bilen_universal_2017,chang_domain-specific_2019,dou_unpaired_2020,karani_lifelong_2018,liu_ms-net_2020}, and a domain-specific output layer. 2) We extend the multi-task, multi-domain segmentation learning framework by integrating a contrastive regularization during optimization. As opposed to classical contrastive approaches that operate on image classes \cite{chen_simple_2020,hadsell_dimensionality_2006,khosla_supervised_2020}, we leverage dataset label information to enhance intra-domain similarity and impose inter-domain margins, and 3) we illustrate the effectiveness of our approach for multi-task, multi-domain segmentation on two sparse, unpaired (from different patient cohorts), and heterogeneous pediatric musculoskeletal MR imaging datasets.

\section{Method}
\label{sec:method}

\begin{figure}[t]
\centering
\begin{adjustbox}{width=\textwidth}
\tikzstyle{dashed}=[dash pattern=on .85pt off .85pt]
\begin{tikzpicture}

%% Borders
\draw[line width=0.1mm, color=darkgray, rounded corners=1] (-.35, .35) -- (-.35,-1.05) -- (3.65,-1.05) -- (3.65,.35) -- cycle;
\draw[line width=0.1mm, color=darkgray, rounded corners=1, dashed] (1.7, .3) -- (1.7,-.3) -- (3.6,-.3) -- (3.6,.3) -- cycle;
\draw[line width=0.1mm, color=darkgray, rounded corners=1, dashed] (1.7, -.4) -- (1.7,-1) -- (3.6,-1) -- (3.6,-.4) -- cycle;

%% Images
\node[inner sep=0pt] (mri_s) at (-.16,-.075)
    {\includegraphics[width=.025\textwidth]{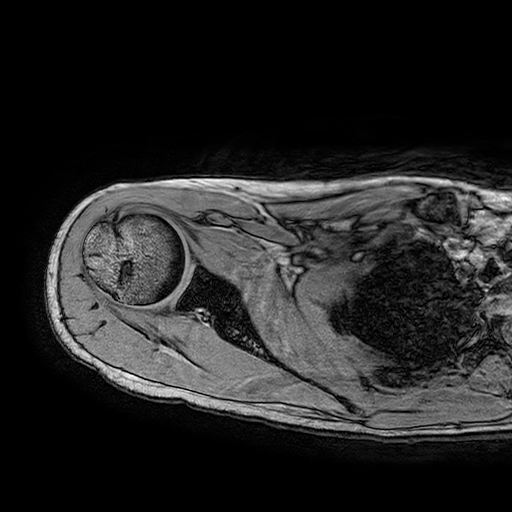}};
\draw [white, rounded corners=1.5, line width=1] (mri_s.north west) -- (mri_s.north east) -- (mri_s.south east) -- (mri_s.south west) -- cycle;
    
\node[inner sep=0pt] (mri_a) at (.16,.075)
    {\includegraphics[width=.025\textwidth]{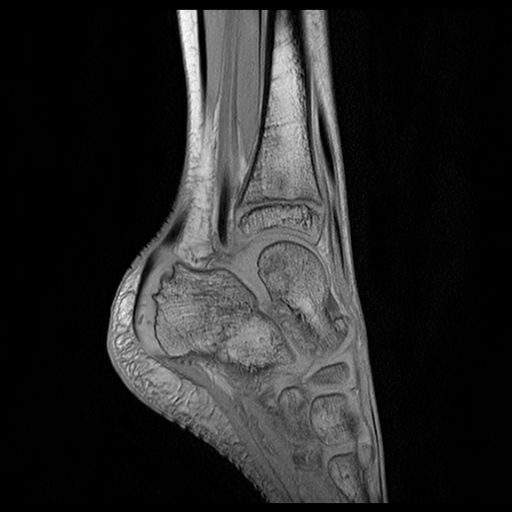}};
\draw [white, rounded corners=1.5, line width=1] (mri_a.north west) -- (mri_a.north east) -- (mri_a.south east) -- (mri_a.south west) -- cycle;

\node[inner sep=0pt] (gt_s) at (1.14,-.075)
    {\includegraphics[width=.025\textwidth]{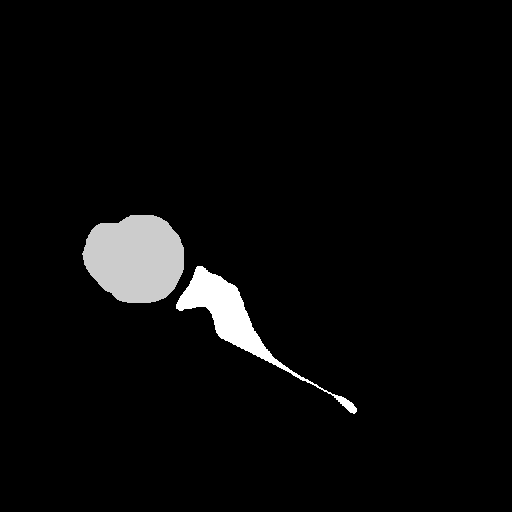}};   
\draw [white, rounded corners=1.5, line width=1] (gt_s.north west) -- (gt_s.north east) -- (gt_s.south east) -- (gt_s.south west) -- cycle;

\node[inner sep=0pt] (gt_a) at (1.46,.075)
    {\includegraphics[width=.025\textwidth]{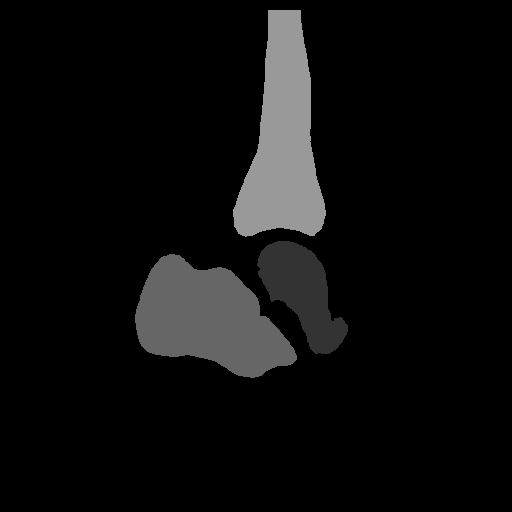}};
\draw [white, rounded corners=1.5, line width=1] (gt_a.north west) -- (gt_a.north east) -- (gt_a.south east) -- (gt_a.south west) -- cycle;
    
%% Legend Images
\node at (0,-.3) {\scalebox{.3}{MRI}};
\node at (1.3,-.3) {\scalebox{.3}{Prediction}};

%% UNet
\draw[line width=0.1mm, fill=BurntOrange!60, rounded corners = 1] (.885,.2229) -- (.65,.125) -- (.35,.25) -- (.35,-.25) -- (.65,-.125) --  (.885,-.2229);
\draw[line width=0.1mm, fill=Lavender!70, rounded corners = 1] (.885,.2229) -- (.95,.25) -- (.95,-.25) --  (.885,-.2229);
\draw[line width=0.01mm, fill=BrickRed!60] (.385,.2354) -- (.415,.2229) -- (.415,-.2229) -- (.385,-.2354) -- cycle;
\draw[line width=0.01mm, pattern={Lines[angle=45,distance={3pt/sqrt(2)}]}, pattern color=black, solid] (.385,.2354) -- (.415,.2229) -- (.415,-.2229) -- (.385,-.2354) -- cycle;
\draw[line width=0.01mm, fill=BrickRed!60] (.485,.1937) -- (.515,.1812) -- (.515,-.1812) -- (.485,-.1937) -- cycle;
\draw[line width=0.01mm, pattern={Lines[angle=45,distance={3pt/sqrt(2)}]}, pattern color=black] (.485,.1937) -- (.515,.1812) -- (.515,-.1812) -- (.485,-.1937) -- cycle;
\draw[line width=0.01mm, fill=BrickRed!60] (.585,.1520) -- (.615,.1395) -- (.615,-.1395) -- (.585,-.1520) -- cycle;
\draw[line width=0.01mm, pattern={Lines[angle=45,distance={3pt/sqrt(2)}]}, pattern color=black] (.585,.1520) -- (.615,.1395) -- (.615,-.1395) -- (.585,-.1520) -- cycle;
\draw[line width=0.01mm, fill=BrickRed!60] (.685,.1395) -- (.715,.1520) -- (.715,-.1520) -- (.685,-.1395) -- cycle;
\draw[line width=0.01mm,  pattern={Lines[angle=45,distance={3pt/sqrt(2)}]}, pattern color=black] (.685,.1395) -- (.715,.1520) -- (.715,-.1520) -- (.685,-.1395) -- cycle;
\draw[line width=0.01mm, fill=BrickRed!60] (.785,.1812) -- (.815,.1937) -- (.815,-.1937) -- (.785,-.1812) -- cycle;
\draw[line width=0.01mm, pattern={Lines[angle=45,distance={3pt/sqrt(2)}]}, pattern color=black] (.785,.1812) -- (.815,.1937) -- (.815,-.1937) -- (.785,-.1812) -- cycle;
\draw[line width=0.01mm, fill=BrickRed!60] (.885,.2229) -- (.915,.2354) -- (.915,-.2354) -- (.885,-.2229) -- cycle;
\draw[line width=0.01mm, pattern={Lines[angle=45,distance={3pt/sqrt(2)}]}, pattern color=black] (.885,.2229) -- (.915,.2354) -- (.915,-.2354) -- (.885,-.2229) -- cycle;
\draw[line width=0.1mm, rounded corners = 1] (.35,.25) -- (.65,.125) -- (.95,.25) -- (.95,-.25) -- (.65,-.125) -- (.35,-.25) -- cycle;
\node at (.65,.275) {\scalebox{.3}{UNet}};

%% Legend UNet
\draw[line width=0.01mm, fill=BurntOrange!60, rounded corners =.5] (-.3,.-.55) -- (-.3,-.6) -- (-.1,-.6) -- (-.1,-.55) -- cycle;
\draw[line width=0.01mm, fill=BrickRed!60, rounded corners =.5] (-.3,.-.7) -- (-.3,-.75) -- (-.1,-.75) -- (-.1,-.7) -- cycle;
\draw[line width=0.01mm, pattern={Lines[angle=45,distance={3pt/sqrt(2)}]}, pattern color=black, rounded corners =.5] (-.3,.-.7) -- (-.3,-.75) -- (-.1,-.75) -- (-.1,-.7) -- cycle;
\draw[line width=0.01mm, fill=Lavender!70, rounded corners =.5] (-.3,.-.85) -- (-.3,-.9) -- (-.1,-.9) -- (-.1,-.85) -- cycle;
\node[anchor=west] at (-.175,-.575) {\scalebox{.3}{Shared convolutional filters}};
\node[anchor=west] at (-.175,-.725) {\scalebox{.3}{Domain-specific batch normalization}};
\node[anchor=west] at (-.175,-.875) {\scalebox{.3}{Domain-specific segmentation layer}};

%% Arrow
\draw[line width=0.1mm, color=darkgray, rounded corners=1, dashed] (.8,.05) -- (.8,.275) -- (1.7,.275);
\draw[line width=0.1mm, color=darkgray] (.8,0) circle (.035);
\draw[line width=0.1mm, color=darkgray, rounded corners=1, dashed] (.65,-.05) -- (.65,-.425) -- (1.7,-.425);
\draw[line width=0.1mm, color=darkgray] (.65,0) circle (.035);

%% Domain-specific batch normalization
\node at (2.65,.225) {\scalebox{.3}{Domain-specific batch normalization}};

%% Graph
\draw[line width=0.05mm, -{Latex[length=1pt, width=1pt]}] (1.875, -.2251) -- (2.725, .0582);
\draw[line width=0.05mm, -{Latex[length=1pt, width=1pt]}] (1.875, -.1751) -- (2.425, -.1751);
\draw[line width=0.05mm, -{Latex[length=1pt, width=1pt]}] (2.1, -.1001) -- (2.65, -.1001);
\draw[line width=0.05mm, -{Latex[length=1pt, width=1pt]}] (2.325, -.0251) -- (2.875, -.0251);
\draw[line width=0.05mm, -{Latex[length=1pt, width=1pt]}] (2.025, -.1751) -- (2.025, -.0001);
\draw[line width=0.05mm, -{Latex[length=1pt, width=1pt]}] (2.25, -.1001) -- (2.25, .0749);
\draw[line width=0.05mm, -{Latex[length=1pt, width=1pt]}] (2.475, -.0251) -- (2.475, .1499);

%% Gaussian density
\draw[line width=0.1mm, domain=-.15:.15, smooth, variable=\x, CornflowerBlue] plot ({\x + 2.5} , {exp(-(\x*\x)/(2*.0036)) * .09 - .0251});
\draw[line width=0.1mm, domain=-.15:.15, smooth, variable=\x, ForestGreen] plot ({\x + 2.6} , {exp(-(\x*\x)/(2*.0025)) * .12 - .0251});
\draw[line width=0.1mm, domain=-.15:.15, smooth, variable=\x, CornflowerBlue] plot ({\x + 2.3} , {exp(-(\x*\x)/(2*.0009)) * .16 - .1001});
\draw[line width=0.1mm, domain=-.15:.15, smooth, variable=\x, ForestGreen] plot ({\x + 2.2} , {exp(-(\x*\x)/(2*.0016)) * .14 - .1001});
\draw[line width=0.1mm, domain=-.15:.15, smooth, variable=\x, CornflowerBlue] plot ({\x + 2.05} , {exp(-(\x*\x)/(2*.0009)) * .15 - .1751});
\draw[line width=0.1mm, domain=-.15:.15, smooth, variable=\x, ForestGreen] plot ({\x + 2.1} , {exp(-(\x*\x)/(2*.0004)) * .17 - .1751});

%% Legend
\node[anchor=west] at (1.925,-.225) {\scalebox{.25}{Features}};
\node at (2.855, .125) {\scalebox{.25}{Network}};
\node at (2.855, .05) {\scalebox{.25}{depth}};
\draw[line width=0.1mm, CornflowerBlue] (3, -.0251) -- (3.15, -.0251);
\node[anchor=west] at (3.05, -.0251) {\scalebox{.3}{Ankle}};
\draw[line width=0.1mm, ForestGreen] (3, -.1751) -- (3.15, -.1751);
\node[anchor=west] at (3.05, -.1751) {\scalebox{.3}{Shoulder}};

%% Shared Representation + Contrastive Loss
\node at (2.65,-.475) {\scalebox{.3}{Shared representation}};

%% Ankle representation
\fill [CornflowerBlue!20, opacity=.75] plot [smooth cycle] coordinates {(1.76, -.71) (1.99,-.58) (2.15, -.7) (2.04, -.87)};
\fill[CornflowerBlue] (1.91,-.75) circle (.02);
\fill[CornflowerBlue] (2.09,-.7) circle (.02);
\fill[CornflowerBlue] (1.84,-.71) circle (.02);
\fill[CornflowerBlue] (2.02,-.79) circle (.02);
\fill[CornflowerBlue] (1.99,-.71) circle (.02);
\fill[CornflowerBlue] (1.93,-.66) circle (.02);
\fill[CornflowerBlue] (2.01,-.66) circle (.02);

%% Shoulder representation
\fill [ForestGreen!20, opacity=.75] plot [smooth cycle] coordinates {(2.06, -.76) (2.24,-.63) (2.38, -.76) (2.29, -.85) (2.16, -.86)};
\fill[ForestGreen] (2.32,-.78) circle (.02);
\fill[ForestGreen] (2.24,-.74) circle (.02);
\fill[ForestGreen] (2.19,-.82) circle (.02);
\fill[ForestGreen] (2.25,-.68) circle (.02);
\fill[ForestGreen] (2.12,-.77) circle (.02);
\fill[ForestGreen] (2.16,-.72) circle (.02);
\fill[ForestGreen] (2.3,-.71) circle (.02);

%% Contrastive loss
\draw[line width=0.15mm, -{Latex[length=2pt, width=2pt]}] (2.43, -.75) -- (2.58, -.75);
\node at (2.505, -.925) {\scalebox{.3}{$\mathcal{L}_{\text{Contrastive}}$}};

%% Ankle representation
\fill [CornflowerBlue!20, opacity=.75] plot [smooth cycle] coordinates {(2.57, -.63) (2.71,-.52) (2.86, -.65) (2.74, -.77)};
\fill[CornflowerBlue] (2.72,-.65) circle (.02);
\fill[CornflowerBlue] (2.76,-.59) circle (.02);
\fill[CornflowerBlue] (2.64,-.63) circle (.02);
\fill[CornflowerBlue] (2.74,-.71) circle (.02);
\fill[CornflowerBlue] (2.71,-.57) circle (.02);
\fill[CornflowerBlue] (2.67,-.69) circle (.02);
\fill[CornflowerBlue] (2.80,-.64) circle (.02);

%% Shoulder representation
\fill [ForestGreen!20, opacity=.75] plot [smooth cycle] coordinates {(2.75, -.84) (2.89,-.71) (3.03, -.82) (2.92, -.96)};
\fill[ForestGreen] (2.9,-.82) circle (.02);
\fill[ForestGreen] (2.81,-.84) circle (.02);
\fill[ForestGreen] (2.93,-.78) circle (.02);
\fill[ForestGreen] (2.86,-.89) circle (.02);
\fill[ForestGreen] (2.96,-.84) circle (.02);
\fill[ForestGreen] (2.84,-.79) circle (.02);
\fill[ForestGreen] (2.92,-.90) circle (.02);

%% Legend
\fill[CornflowerBlue] (3.1,-.7251) circle (.02);
\node[anchor=west] at (3.05, -.7251) {\scalebox{.3}{Ankle}};
\fill[ForestGreen] (3.1,-.8751) circle (.02);
\node[anchor=west] at (3.05, -.8751) {\scalebox{.3}{Shoulder}};

\end{tikzpicture}
\end{adjustbox}
  \caption{The proposed multi-task, multi-domain segmentation method is based on UNet \cite{ronneberger_u-net_2015} with shared convolutional filters as well as domain-specific batch normalization and a domain-specific segmentation layer. The training procedure incorporates a contrastive loss $\mathcal{L}_{\text{Contrastive}}$ to promote inter-domain separation in the shared representation.}
  \label{fig:proposed_method}
\end{figure}
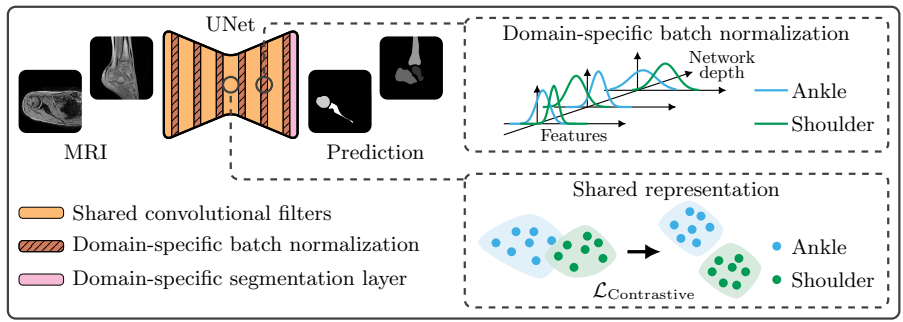

\subsection{Deep Segmentation Model with Domain-Specific Layers (DSL)}
\label{sec:deep_segmentation_model_with_domain_specific_layers}

Let $\mathcal{D}_{1}, ..., \mathcal{D}_{K}$ be $K$ different datasets organized such that the $k$-th dataset $\mathcal{D}_k = \{x_{i}^{k}, y_{i}^{k}\}_{i=1}^{n_{k}}$ contains $n_k$ pairs of greyscale images $x_{i}^{k}$ and corresponding class label images $y_{i}^{k}$ in label space $\mathscr{C}_{k}$. We considered a network $S: x_{i}^{k} \mapsto S(x_{i}^{k}; \Theta, \Lambda_{k}, W_{k})$ with shared parameters $\Theta$ and domain-specific weights $\{\Lambda_{k}, W_{k}\}_{k=1}^{K}$. The shared parameters $\Theta$ comprised classical and transposed convolutional filters, while $\Lambda_{k}$ represented domain-specific batch normalization weights and $W_{k}$ corresponded to the weights of the domain-specific segmentation layer (Fig. \ref{fig:proposed_method}). 

Batch normalization aims at improving convergence speed and generalization abilities of the model by normalizing the internal activations of the network \cite{ioffe_batch_2015}. However, as the individual statistics of the $K$ domains can be very different from each other, a domain-agnostic batch normalization layer could lead to defective features. This could result in zero mean activation over domains at certain layers which is meaningless \cite{bilen_universal_2017,chang_domain-specific_2019,dou_unpaired_2020,karani_lifelong_2018,liu_ms-net_2020}. Thus, to more carefully calibrate the internal activations, we employed domain-specific batch normalization ($\text{DSBN}$) functions:
\begin{equation}
    \text{DSBN}(v_{l,m}^{k}; \beta_{l,m}^{k}, \gamma_{l,m}^{k}) = \gamma_{l,m}^{k} \dfrac{v_{l,m}^{k} - \mathbb{E}[v_{l,m}^{k}]}{\sqrt{\mathbb{V}[v_{l,m}^{k}] + \epsilon}}  + \beta_{l,m}^{k}
\end{equation}
\noindent where $v_{l,m}^{k}$ denoted the $m$-th features at the $l$-th layer produced by an input batch from the $k$-th dataset and $\epsilon = 1\text{e-}5$ was added for numerical stability. Hence, $\Lambda_{k} = \{\beta_{l,m}^{k}, \gamma_{l,m}^{k}\}_{l,m}$ comprised the domain-specific trainable shift and scale of each features at each layers.

\begin{figure}[t]
\centering
\begin{adjustbox}{width=\textwidth}
\tikzstyle{dashed}=[dash pattern=on 10pt off 10pt]
\begin{tikzpicture}

%% UNet
\draw[line width=0.01mm, fill=black] (-.1,-2) rectangle (0,2);
\draw[line width=0.01mm, fill=black] (1,-2) rectangle (1.5,2);

\draw[line width=0.01mm, fill=black] (2.5,-1) rectangle (3,1);
\draw[line width=0.01mm, fill=black] (4,-1) rectangle (4.7,1);

\draw[line width=0.01mm, fill=black] (5.7,-.5) rectangle (6.4,.5);
\draw[line width=0.01mm, fill=black] (7.4,-.5) rectangle (8.5,.5);

\draw[line width=0.01mm, fill=black] (9.5,-0.25) rectangle (10.6,0.25);
\draw[line width=0.01mm, fill=black] (11.6,-0.25) rectangle (13.3,0.25);

\draw[line width=0.01mm, fill=black] (14.3,-0.125) rectangle (16,0.125);
\draw[line width=0.01mm, fill=black] (17,-0.125) rectangle (19.5,0.125);

\draw[line width=0.01mm, fill=black] (20.5,-.25) rectangle (22.2,.25);
\draw[line width=0.01mm] (22.2,-.25) rectangle (23.9,.25);
\draw[line width=0.01mm, fill=black] (24.9,-.25) rectangle (26.6,.25);

\draw[line width=0.01mm, fill=black] (27.6,-.5) rectangle (28.7,.5);
\draw[line width=0.01mm] (28.7,-.5) rectangle (29.8,.5);
\draw[line width=0.01mm, fill=black] (30.8,-.5) rectangle (31.9,.5);

\draw[line width=0.01mm, fill=black] (32.9,-1) rectangle (33.6,1);
\draw[line width=0.01mm] (33.6,-1) rectangle (34.3,1);
\draw[line width=0.01mm, fill=black] (35.3,-1) rectangle (36,1);

\draw[line width=0.01mm, fill=black] (37,-2) rectangle (37.5,2);
\draw[line width=0.01mm] (37.5,-2) rectangle (38,2);
\draw[line width=0.01mm, fill=black] (39,-2) rectangle (39.5,2);
\draw[line width=0.01mm, fill=black] (40.5,-2) rectangle (40.6,2);

%% Skip connections
\draw[line width=1mm, -{Latex[length=15pt, width=15pt]}] (1.25,2) -- (1.25,2.35) -- (19.5,2.35);
\draw[line width=1mm] (19.5,2.35) -- (37.75,2.35) -- (37.75,2);
\draw[line width=1mm, -{Latex[length=15pt, width=15pt]}] (4.35,1) -- (4.35,1.35) -- (19.2,1.35);
\draw[line width=1mm] (19.2,1.35) -- (34.05,1.35) -- (34.05,1);
\draw[line width=1mm, -{Latex[length=15pt, width=15pt]}] (7.95,.5) -- (7.95,.85) -- (18.6,.85);
\draw[line width=1mm] (18.6,.85) -- (29.25,.85) -- (29.25,.5);
\draw[line width=1mm, -{Latex[length=15pt, width=15pt]}] (12.45,.25) -- (12.45,.6) -- (17.75,.6);
\draw[line width=1mm] (17.75,.6) -- (23.05,.6) -- (23.05,.25);

%% Arrow
\draw[line width=0.01mm, fill=ForestGreen] (.2,.4) -- (.2,-.4) -- (.8,0) -- cycle;
\draw[line width=0.01mm, fill=BrickRed] (1.7,.4) -- (1.7,-.4) -- (2.3,0) -- cycle;

\draw[line width=0.01mm, fill=ForestGreen] (3.2,.4) -- (3.2,-.4) -- (3.8,0) -- cycle;
\draw[line width=0.01mm, fill=BrickRed] (4.9,.4) -- (4.9,-.4) -- (5.5,0) -- cycle;

\draw[line width=0.01mm, fill=ForestGreen] (6.6,.4) -- (6.6,-.4) -- (7.2,0) -- cycle;
\draw[line width=0.01mm, fill=BrickRed] (8.7,.4) -- (8.7,-.4) -- (9.3,0) -- cycle;

\draw[line width=0.01mm, fill=ForestGreen] (10.8,.4) -- (10.8,-.4) -- (11.4,0) -- cycle;
\draw[line width=0.01mm, fill=BrickRed] (13.5,.4) -- (13.5,-.4) -- (14.1,0) -- cycle;

\draw[line width=0.01mm, fill=ForestGreen] (16.2,.4) -- (16.2,-.4) -- (16.8,0) -- cycle;
\draw[line width=0.01mm, fill=CornflowerBlue] (19.7,.4) -- (19.7,-.4) -- (20.3,0) -- cycle;

\draw[line width=0.01mm, fill=ForestGreen] (24.1,.4) -- (24.1,-.4) -- (24.7,0) -- cycle;
\draw[line width=0.01mm, fill=CornflowerBlue] (26.8,.4) -- (26.8,-.4) -- (27.4,0) -- cycle;

\draw[line width=0.01mm, fill=ForestGreen] (30,.4) -- (30,-.4) -- (30.6,0) -- cycle;
\draw[line width=0.01mm, fill=CornflowerBlue] (32.1,.4) -- (32.1,-.4) -- (32.7,0) -- cycle;

\draw[line width=0.01mm, fill=ForestGreen] (34.5,.4) -- (34.5,-.4) -- (35.1,0) -- cycle;
\draw[line width=0.01mm, fill=CornflowerBlue] (36.2,.4) -- (36.2,-.4) -- (36.8,0) -- cycle;

\draw[line width=0.01mm, fill=ForestGreen] (38.2,.4) -- (38.2,-.4) -- (38.8,0) -- cycle;
\draw[line width=0.01mm, fill=Lavender] (39.7,.4) -- (39.7,-.4) -- (40.3,0) -- cycle;

\draw[line width=0.01mm, fill=BurntOrange] (17.85,-1.45) -- (18.65,-1.45) -- (18.25,-2.05) -- cycle;

%% Text
\node[rotate=90] at (-0.7,0) {\scalebox{3.25}{256$\times$256}};
\node at (-0.05,-2.6) {\scalebox{3.25}{1}};
\node at (1.25,-2.6) {\scalebox{3.25}{32}};

\node at (2.75,-1.6) {\scalebox{3.25}{32}};
\node at (4.35,-1.6) {\scalebox{3.25}{64}};

\node at (6.05,-1.1) {\scalebox{3.25}{64}};
\node at (7.95,-1.1) {\scalebox{3.25}{128}};

\node at (10.05,-.85) {\scalebox{3.25}{128}};
\node at (12.45,-.85) {\scalebox{3.25}{256}};

\node at (15.15,-.725) {\scalebox{3.25}{256}};
\node at (18.25,-.725) {\scalebox{3.25}{512}};

\node at (21.35,-.85) {\scalebox{3.25}{256}};
\node at (23.05,-.85) {\scalebox{3.25}{256}};
\node at (25.75,-.85) {\scalebox{3.25}{256}};

\node at (27.9,-1.1) {\scalebox{3.25}{128}};
\node at (29.5,-1.1) {\scalebox{3.25}{128}};
\node at (31.35,-1.1) {\scalebox{3.25}{128}};

\node at (33.05,-1.6) {\scalebox{3.25}{64}};
\node at (34.15,-1.6) {\scalebox{3.25}{64}};
\node at (35.65,-1.6) {\scalebox{3.25}{64}};

\node at (36.95,-2.6) {\scalebox{3.25}{32}};
\node at (38.05,-2.6) {\scalebox{3.25}{32}};
\node at (39.25,-2.6) {\scalebox{3.25}{32}};
\node at (40.55,-2.6) {\scalebox{3.25}{$\vert\mathscr{C}_{k}\vert$}};

\node at (18.25,-2.6) {\scalebox{3.25}{$z$}};

%% Network name
\node at (20.25,3.2) {\scalebox{3.25}{Segmentation network $S$ based on UNet}};

% Legend
\draw[dashed, line width=1mm, color=darkgray, rounded corners=10] (1.9,-3.5) rectangle (38.6,-8.5);

\draw[line width=0.01mm, fill=ForestGreen] (2.9,-4.35) -- (2.9,-5.15) -- (3.5,-4.75) -- cycle;
\draw[line width=0.01mm, fill=BrickRed] (2.9,-5.6) -- (2.9,-6.4) -- (3.5,-6) -- cycle;
\draw[line width=0.01mm, fill=CornflowerBlue] (2.9,-6.85) -- (2.9,-7.65) -- (3.5,-7.25) -- cycle;
\draw[line width=0.01mm, fill=Lavender] (20.25,-5.6) -- (20.25,-6.4) -- (20.85,-6) -- cycle;
\draw[line width=0.01mm, fill=BurntOrange] (20.25,-6.85) -- (20.25,-7.65) -- (20.85,-7.25) -- cycle;

\node[anchor=west] at (4,-4.75) {\scalebox{3.25}{Shared double conv 3$\times$3, domain-specific batch normalization, ReLU}};
\node[anchor=west] at (4,-6) {\scalebox{3.25}{Max-pooling 2$\times$2}};
\node[anchor=west] at (4,-7.25) {\scalebox{3.25}{Shared transpose conv 2$\times$2}};
\node[anchor=west] at (21.25,-6) {\scalebox{3.25}{Domain-specific conv 1$\times$1, softmax}};
\node[anchor=west] at (21.25,-7.25) {\scalebox{3.25}{Global max pooling}};

\end{tikzpicture}
\end{adjustbox}
\caption{The architecture of the segmentation network $S$ is based on a convolutional encoder-decoder and a final segmentation layers with $\vert \mathscr{C}_{k} \vert$ classes. The image embedding $z$ is obtained via UNet encoder and global max pooling.}
\label{fig:proposed_architecture}
\end{figure}
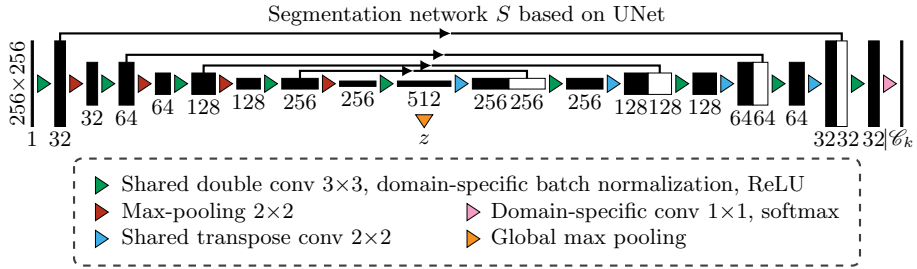

Moreover, it is essential to employ a dedicated output layer, as a domain-agnostic segmentation layer may predict classes from all $K$ datasets, which is counterproductive \cite{fourure_semantic_2016}. Hence, the network was designed such that $S = \psi \circ \phi$ where $\psi$ was the segmentation layer while $\phi$ contained all the previous layers of the network. Specifically, if $u_{i}^{k} = \phi(x_{i}^{k}; \Theta, \Lambda_{k})$ denotes the output of the penultimate layer then $\psi(u_{i}^{k}; W_{k}) = \text{softmax}(u_{i}^{k} * W_{k})$ was a domain-specific segmentation layer with a 1$\times$1 convolutional filter $W_{k}$ which produced a segmentation mask with $\vert \mathscr{C}_{k} \vert$ classes (Fig. \ref{fig:proposed_architecture}). $\vert \mathscr{C}_{k} \vert$ denoted the cardinality of the $k$-th label space.

During training, we used the stochastic gradient descent algorithm to optimize the cross-entropy loss defined in a multi-task and multi-domain setting:
\begin{equation}
    \mathcal{L}_{\text{CE}} = - \dfrac{1}{K} \sum_{k=1}^{K} \dfrac{1}{n_{k} \vert \mathscr{C}_{k} \vert} \sum_{i=1}^{n_{k}} \sum_{c \in \mathscr{C}_{k}} y_{i,c}^{k} \log(\hat{y}_{i,c}^{k})
\end{equation}
\noindent where $\hat{y}_{i}^{k} = S(x_{i}^{k}; \Theta, \Lambda_{k}, W_{k})$ was the predicted segmentation. The shared parameters and the domain-specific weights were thus learnt through this novel optimization scheme. The domain-specific weights were a minimal supplementary parameterization with regards to the number of shared convolutional filters.

\subsection{Supervised Contrastive Regularization}
\label{sec:supervised_contrastive_regularization}

The encoder of the segmentation network mapped the greyscale images to a shared representation (Fig. \ref{fig:proposed_method}). We assumed that learning a shared representation with domain-specific clusters would enhance the generalizability of the decoder and improve the accuracy of the segmentation predictions. More precisely, we assumed that a local variation in the shared representation should preserve the category of the domain. Hence, we designed a novel regularization term aimed at conserving intra-domain cohesion and inter-domain separation in the shared representation. We adapted the supervised contrastive loss \cite{khosla_supervised_2020} to multi-domain segmentation using the known datasets labels. In the following, $x_{i}$ denotes the $i$-th image of the joint dataset with $n$ samples.

Let $z_{i}$ be the embedding of $x_{i}$ to which we applied global max pooling to obtain a spatially invariant representation (Fig. \ref{fig:proposed_architecture}), and $k_{i}$ the dataset index of $x_{i}$. We note $P(i) = \{p \in [\![1,..., n ]\!]: k_{p} = k_{i}, p \neq i\}$ the set of indexes of all images from the same dataset as $x_i$. As opposed to classical self-supervised contrastive approaches \cite{chen_simple_2020,hadsell_dimensionality_2006,khosla_supervised_2020}, it was not necessary to generate an augmented view of $x_{i}$ as more than one sample was known to be from the same dataset. The contrastive loss was defined as follows:
\begin{equation}
    \mathcal{L}_{\text{Contrastive}} = - \frac{1}{n} \sum_{i=1}^{n} \frac{1}{|P(i)|} \sum_{p \in P(i)} \log \left( \frac{\exp(\text{sim}(z_i, z_p) / \tau)}{\sum_{j=1}^{n} \mathbb{1}_{[j \neq i]} \exp(\text{sim}(z_i, z_j) / \tau)} \right) 
\end{equation}
\noindent where $\text{sim}(z_{i}, z_{j}) = z_{i} \cdot z_{j} / \|z_{i}\| \|z_{j}\|$ was the cosine similarity between two representations and $\tau$ was the temperature hyper-parameter which controlled the smoothness of the loss as well as imposed hard negative/positive predictions \cite{chen_simple_2020,khosla_supervised_2020}. During optimization, only the weights of the encoder were penalized by the proposed regularization. Thus, the contrastive regularization gathered the representation from the same domain in embedded space, while simultaneously separating clusters from different domains (Fig. \ref{fig:proposed_method}). The deep segmentation model with domain-specific
layers (DSL) was trained using the proposed regularized loss $\mathcal{L} = \mathcal{L}_{\text{CE}} + \lambda \mathcal{L}_{\text{Contrastive}}$ with weighting hyper-parameter $\lambda$.

\section{Experiments}
\label{sec:experiments}

\subsection{Imaging Datasets}
\label{sec:imaging_datasets}

Experiments were conducted on MR images retrospectively extracted from previous clinical studies. These image datasets were acquired using a 3T Achieva scanner (Philips Healthcare, Best, Netherlands) from two pediatric musculoskeletal joints: ankle and shoulder. The ankle dataset was acquired from 17 pediatric patients aged from 7 to 13 years ($10\pm2$ years) while MR images of 15 shoulder were acquired from pediatric patients aged from 5 to 17 years old ($12\pm4$ years). The ankle dataset comprised 10 healthy and 7 pathological cases and the shoulder dataset comprised 8 healthy and 7 pathological cases. A T1 weighted 3D gradient echo sequence was used (TR: 7.9 ms, TE: 2.8 ms, voxel: $0.4\times0.4\times1.2$ mm$^{3}$, FOV: $140\times161$ mm$^{2}$) to acquire ankle images. An eTHRIVE (enhanced T1-weighted High-Resolution Isotropic Volume Examination) sequence was employed (TR: 8.4 ms, TE: 4.2 ms, voxel: $0.4\times0.4\times1.2$ mm$^{3}$, FOV: $2600\times210$ mm$^{2}$) for shoulder image acquisition. Magnetic resonance images were acquired during separate clinical studies where parents provided informed consent to use the imaging data for research purpose. A medically trained expert (12 years of experience) annotated the images to obtain ground truth segmentation masks of calcaneus, talus and tibia bones for ankle and scapula and humerus bones for shoulder. All axial slices were downsampled to $256\times256$ pixels and were normalized to have zero-mean and unit variance.

\subsection{Implementation Details}
\label{sec:implementation_details}

The proposed method based on DSL and contrastive regularization was implemented using UNet architecture \cite{ronneberger_u-net_2015} and its performance was assessed in comparison with other state-of-the-art models. The compared methods included UNet \cite{ronneberger_u-net_2015} and UNet with attention gates (Att-UNet) \cite{oktay_attention_2018}, which were implemented using baseline (trained on individual dataset), joint (trained on all datasets at once) and DSL schemes. All networks were initialized with randomly distributed weights and trained from scratch without any transfer learning scheme.

All methods were based on the same backbone UNet architecture (Fig. \ref{fig:proposed_architecture}). The segmentation networks were trained for 30 epochs with batch size set to 32, using the Adam optimizer with a $1\text{e-}4$ learning rate. We explored different values for the hyper-parameters of the supervised contrastive regularization and found $\tau = 1\text{e-}1$ and $\lambda = 1\text{e-}1$ to be optimal. We used extensive on the fly 2D data augmentation including random flip, translation ($\pm25\%$) and rotation ($\pm45^{\circ}$) in both directions due to limited available training data. Deep learning networks were implemented on PyTorch and a Nvidia RTX 2080 Ti GPU with 12 GB of RAM was used during optimization.

The same post-processing was employed after each method: first, the obtained 2D segmentation masks were stacked to form a 3D binary volume, then we selected the 3D largest connected set as output prediction and we finally applied 3D morphological closing ($5\times5\times5$ spherical kernel) to smooth the resulting boundaries.

\subsection{Evaluation of Predicted Segmentation}
\label{sec:evaluation_of_predicted_segmentation}

We assessed the performance of each method based on the similarity between 3D predicted and ground truth masks. For each dataset, Dice coefficient, average symmetric surface distance (ASSD) and maximum symmetric surface distance (MSSD) metrics were computed for each bone and we reported the average scores. Due to the limited amount of examinations, the metrics were determined in a leave-one-out manner such that, for each dataset, one examination was retained for validation, one for test and the remaining data were used to train the model. We iterated through the datasets simultaneously and each examination was used at maximum once for test. We did not test all combinations between datasets, as this would have drastically increased computation time and would have introduced redundant observations in the results. Moreover, due to the scarce amount of 3D examinations, we performed statistical analysis between the methods on the 2D MR images from both datasets. We employed the Kolmogorov-Smirnov non-parametric test using Dice scores obtained from the $6215$ ankle and shoulder 2D slices corresponding to 32 3D MR images.

\section{Results and Discussion}
\label{sec:results_and_discussion}

\begin{table}[t]
\caption{Quantitative assessment of UNet \cite{ronneberger_u-net_2015} and Att-UNet \cite{oktay_attention_2018} using baseline, joint and DSL schemes, and the proposed UNet with DSL and contrastive regularization on ankle and shoulder datasets. Metrics include Dice ($\%$), ASSD (mm) and MSSD (mm). Best results are in bold.}
\centering
    \begin{tabular}{|P{.45cm}|P{1cm}||P{1.6cm}|P{1.6cm}|P{1.6cm}|P{1.6cm}|P{1.6cm}|P{1.6cm}|P{1.6cm}|P{1.6cm}|} 
    \hline
    \multicolumn{2}{|c||}{\multirow{2}{*}{Metric}} & \multicolumn{3}{c|}{Ankle} & \multicolumn{3}{c|}{Shoulder} \\\cline{3-8}
    \multicolumn{2}{|c||}{} & Dice $\uparrow$ & ASSD $\downarrow$ & MSSD $\downarrow$ & Dice $\uparrow$  & ASSD $\downarrow$ & MSSD $\downarrow$\\ 
    \hline\hline
     
    \multirow{3}{*}{\rotatebox[origin=c]{90}{UNet}} & Base & 92.2$\pm$3.2 & 1.0$\pm$1.1 & 9.5$\pm$9.0 & 81.8$\pm$13.8  & 2.6$\pm$3.8 & 22.3$\pm$15.1 \\\cline{2-8}
    & Joint & 93.0$\pm$3.3 & \textbf{0.8$\pm$0.6} & 8.5$\pm$8.0 & 83.6$\pm$11.0 & 2.1$\pm$2.6 & 21.9$\pm$19.3 \\\cline{2-8}
    & DSL & 93.4$\pm$1.5 & \textbf{0.8$\pm$0.5} & 8.3$\pm$7.8 & 84.3$\pm$10.7 & 2.2$\pm$2.4 & 25.4$\pm$26.4 \\\hline
    
    \multirow{3}{*}{\rotatebox[origin=c]{90}{Att}} & Base & 92.4$\pm$2.0 & 1.0$\pm$1.1 & 8.9$\pm$7.4 & 83.0$\pm$13.1 & 2.0$\pm$3.5 & 19.5$\pm$16.7 \\\cline{2-8}
    & Joint & 92.7$\pm$1.6 & 0.9$\pm$0.7 & 9.4$\pm$7.5 & 83.2$\pm$13.0 & 2.3$\pm$3.9 & 20.2$\pm$18.3 \\\cline{2-8}
    & DSL & 93.4$\pm$1.6 & \textbf{0.8$\pm$0.4} & 8.8$\pm$7.9 & 83.8$\pm$12.1 & 2.2$\pm$4.1 & 18.4$\pm$15.6 \\\hline\hline
  
    \multicolumn{2}{|c||}{Proposed} & \textbf{93.5$\pm$1.2} & \textbf{0.8$\pm$0.5} & \textbf{7.8$\pm$7.8} & \textbf{85.8$\pm$8.6} & \textbf{1.4$\pm$1.4} & \textbf{18.3$\pm$11.1} \\\hline
    
    \end{tabular}
\label{tab:results}
\end{table}

\begin{table}[t]
\caption{Statistical analysis between the proposed model with UNet \cite{ronneberger_u-net_2015} and Att-UNet \cite{oktay_attention_2018} using baseline, joint and DSL schemes, through Kolmogorov-Smirnov non-parametric test using Dice computed on 2D slices from ankle and shoulder datasets. Bold \textit{p}-values ($<0.05$) highlight statistically significant results.}
\centering
    \begin{tabular}{|P{1.25cm}||P{1.4cm}|P{1.4cm}|P{1.4cm}|P{1.4cm}|P{1.4cm}|P{1.4cm}|P{1.4cm}|} 
    \hline
    \multirow{2}{*}{Method} & \multicolumn{3}{c|}{UNet} & \multicolumn{3}{c|}{Att-UNet} & \multirow{2}{*}{Proposed} \\\cline{2-7}
     & Base & Joint & DSL & Base & Joint & DSL & \\\hline\hline
    
    Dice 2D & 86.9$\pm$18.7 & 87.6$\pm$18.1 & 88.3$\pm$17.6 & 87.8$\pm$18.5 & 87.5$\pm$18.3 & 88.5$\pm$16.6 & 88.9$\pm$17.0 \\\hline
    \textit{p}-value & \textbf{1.5\text{e-}15} & \textbf{3.5\text{e-}6} & \textbf{0.04} & \textbf{7.2\text{e-}3} & \textbf{2.6\text{e-}8} & \textbf{1.0\text{e-}4} & \--- \\\hline
    
    \end{tabular}
\label{tab:statistical_analsyis}
\end{table}

\begin{figure}[ht!]
\centering
\begin{adjustbox}{width=\textwidth}
\begin{tikzpicture}
\begin{scope}[spy using outlines=
      {circle, magnification=2.5, size=.36cm, connect spies, rounded corners}]

%% Pictures
%% Ankle 
\node[inner sep=0pt] at (0,0)
    {\includegraphics[width=.115\textwidth]{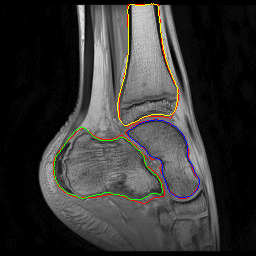}};
\node[inner sep=0pt] at (1.5,0)
    {\includegraphics[width=.115\textwidth]{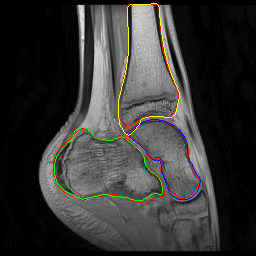}};
\node[inner sep=0pt] at (3,0)
    {\includegraphics[width=.115\textwidth]{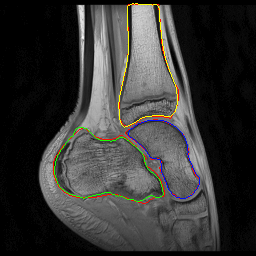}}; 
\node[inner sep=0pt] at (4.5,0)
    {\includegraphics[width=.115\textwidth]{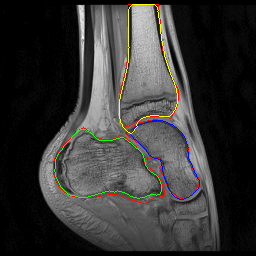}};
\node[inner sep=0pt] at (6,0)
    {\includegraphics[width=.115\textwidth]{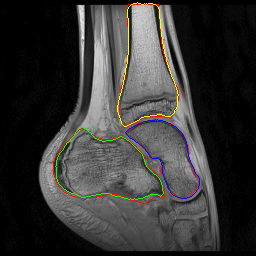}};
\node[inner sep=0pt] at (7.5,0)
    {\includegraphics[width=.115\textwidth]{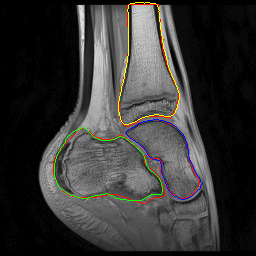}};
\node[inner sep=0pt] at (9,0)
    {\includegraphics[width=.115\textwidth]{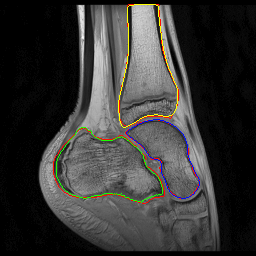}};
    
%% Shoulder
\node[inner sep=0pt] at (0,-1.5)
    {\includegraphics[width=.115\textwidth]{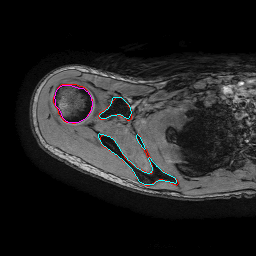}};
\node[inner sep=0pt] at (1.5,-1.5)
    {\includegraphics[width=.115\textwidth]{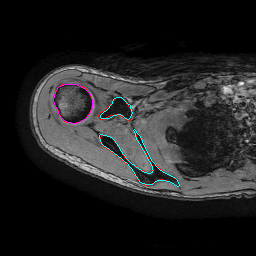}};
    \node[inner sep=0pt] at (3,-1.5)
{\includegraphics[width=.115\textwidth]{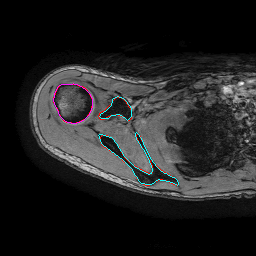}};
\node[inner sep=0pt] at (4.5,-1.5)
    {\includegraphics[width=.115\textwidth]{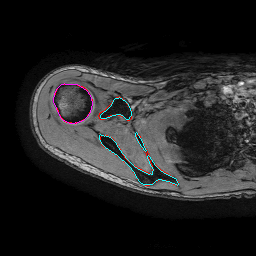}};
\node[inner sep=0pt] at (6,-1.5)
    {\includegraphics[width=.115\textwidth]{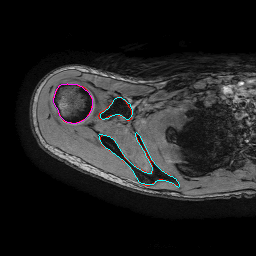}};
\node[inner sep=0pt] at (7.5,-1.5)
    {\includegraphics[width=.115\textwidth]{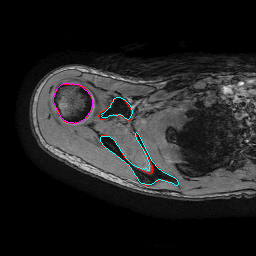}};
\node[inner sep=0pt] at (9,-1.5)
    {\includegraphics[width=.115\textwidth]{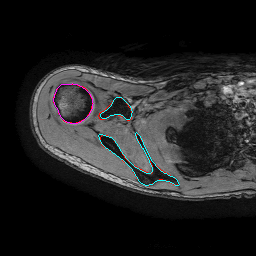}};
    
%% Zoom
%% Ankle 1
\spy [Dandelion] on (0,.02) in node [left] at (-.3,.48);
\spy [Dandelion] on (.14,-.21) in node [left] at (.66,.48);
\spy [Dandelion] on (.11,-.38) in node [left] at (-.3,.-.48);

%% Ankle 2
\spy [Dandelion] on (1.5,.02) in node [left] at (1.2,.48);
\spy [Dandelion] on (1.64,-.21) in node [left] at (2.16,.48);
\spy [Dandelion] on (1.61,-.38) in node [left] at (1.2,.-.48);

%% Ankle 3
\spy [Dandelion] on (3,.02) in node [left] at (2.7,.48);
\spy [Dandelion] on (3.14,-.21) in node [left] at (3.66,.48);
\spy [Dandelion] on (3.11,-.38) in node [left] at (2.7,.-.48);

%% Ankle 4
\spy [Dandelion] on (4.5,.02) in node [left] at (4.2,.48);
\spy [Dandelion] on (4.64,-.21) in node [left] at (5.16,.48);
\spy [Dandelion] on (4.61,-.38) in node [left] at (4.2,.-.48);

%% Ankle 5
\spy [Dandelion] on (6,.02) in node [left] at (5.7,.48);
\spy [Dandelion] on (6.14,-.21) in node [left] at (6.66,.48);
\spy [Dandelion] on (6.11,-.38) in node [left] at (5.7,.-.48);

%% Ankle 6
\spy [Dandelion] on (7.5,.02) in node [left] at (7.2,.48);
\spy [Dandelion] on (7.64,-.21) in node [left] at (8.16,.48);
\spy [Dandelion] on (7.61,-.38) in node [left] at (7.2,.-.48);

%% Ankle 7
\spy [Dandelion] on (9,.02) in node [left] at (8.7,.48);
\spy [Dandelion] on (9.14,-.21) in node [left] at (9.66,.48);
\spy [Dandelion] on (9.11,-.38) in node [left] at (8.7,.-.48);

%% Shoulder 1
\spy [Dandelion] on (-.38,-1.37) in node [left] at (-.3,-1.02);
\spy [Dandelion] on (0,-1.41) in node [left] at (.66,-1.02);
\spy [Dandelion] on (.11,-1.64) in node [left] at (.66,.-1.98);

%% Shoulder 2
\spy [Dandelion] on (1.12,-1.37) in node [left] at (1.2,-1.02);
\spy [Dandelion] on (1.5,-1.41) in node [left] at (2.16,-1.02);
\spy [Dandelion] on (1.61,-1.64) in node [left] at (2.16,.-1.98);

%% Shoulder 3
\spy [Dandelion] on (2.62,-1.37) in node [left] at (2.7,-1.02);
\spy [Dandelion] on (3,-1.41) in node [left] at (3.66,-1.02);
\spy [Dandelion] on (3.11,-1.64) in node [left] at (3.66,.-1.98);

%% Shoulder 4
\spy [Dandelion] on (4.12,-1.37) in node [left] at (4.2,-1.02);
\spy [Dandelion] on (4.5,-1.41) in node [left] at (5.16,-1.02);
\spy [Dandelion] on (4.61,-1.64) in node [left] at (5.16,.-1.98);

%% Shoulder 5
\spy [Dandelion] on (5.62,-1.37) in node [left] at (5.7,-1.02);
\spy [Dandelion] on (6,-1.41) in node [left] at (6.66,-1.02);
\spy [Dandelion] on (6.11,-1.64) in node [left] at (6.66,.-1.98);

%% Shoulder 6
\spy [Dandelion] on (7.12,-1.37) in node [left] at (7.2,-1.02);
\spy [Dandelion] on (7.5,-1.41) in node [left] at (8.16,-1.02);
\spy [Dandelion] on (7.61,-1.64) in node [left] at (8.16,.-1.98);

%% Shoulder 7
\spy [Dandelion] on (8.62,-1.37) in node [left] at (8.7,-1.02);
\spy [Dandelion] on (9,-1.41) in node [left] at (9.66,-1.02);
\spy [Dandelion] on (9.11,-1.64) in node [left] at (9.66,.-1.98);

\end{scope}
    
%% Legend
\node at (0, 1.05) {\scalebox{.7}{UNet}};
\node at (0, .825) {\scalebox{.7}{Base}};
\node at (1.5, 1.05) {\scalebox{.7}{UNet}};
\node at (1.5, .825) {\scalebox{.7}{Joint}};
\node at (3, 1.05) {\scalebox{.7}{UNet}};
\node at (3, .825) {\scalebox{.7}{DSL}};
\node at (4.5, 1.05) {\scalebox{.7}{Att-UNet}};
\node at (4.5, .825) {\scalebox{.7}{Base}};
\node at (6, 1.05) {\scalebox{.7}{Att-UNet}};
\node at (6, .825) {\scalebox{.7}{Joint}};
\node at (7.5, 1.05) {\scalebox{.7}{Att-UNet}};
\node at (7.5, .825) {\scalebox{.7}{DSL}};
\node at (9, .825) {\scalebox{.7}{Proposed}};

\end{tikzpicture}
\end{adjustbox}
\caption{Visual comparison of UNet \cite{ronneberger_u-net_2015} and Att-UNet \cite{oktay_attention_2018} using baseline, joint and DSL schemes, and the proposed model on ankle and shoulder datasets. Ground truth delineations are in red (\textcolor{red}{\---}). Predicted calcaneus, talus, tibia, humerus and scapula bones respectively appear in green (\textcolor{green}{\---}), blue (\textcolor{blue}{\---}), yellow (\textcolor{yellow}{\---}), magenta (\textcolor{magenta}{\---}) and cyan (\textcolor{cyan}{\---}).}
\label{fig:visual_comparison}
\end{figure}
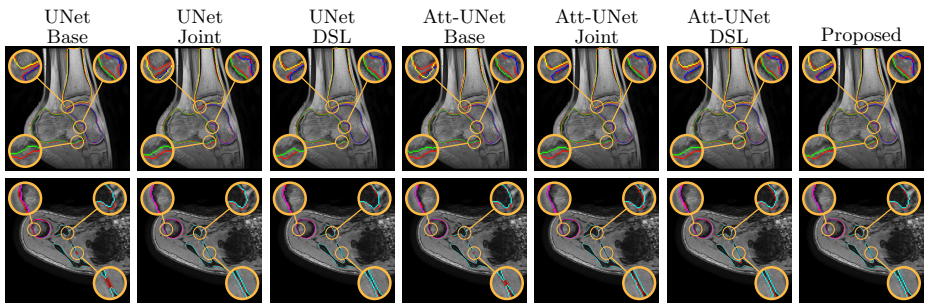

\subsection{Segmentation Results}
\label{sec:segementation_results}

From the quantitative results (Table \ref{tab:results}), the proposed method ranked first on all metrics on both datasets. Results obtained on the ankle dataset corresponded to a marginal increase in performance compared to the second best method ($+0.1\%$ Dice and $-0.5$ mm MSSD) while shoulder performance were substantially increased ($+1.5\%$ Dice, $-0.6$ mm ASSD and $-0.1$ mm MSSD). We observed that for a fixed architecture, the DSL scheme outperformed the joint approach, which in turn outranked the baseline scheme. Hence, the shared representation and layer specialization allowed performance improvements over independent models and promoted more precise bone extraction. We also reported a higher variability in shoulder results, which was due to the presence of examinations with a higher level of noise due to patient movements during acquisition. Moreover, the statistical analysis performed on 2D slices using Dice (Table \ref{tab:statistical_analsyis}) indicated that the proposed model produced significant improvements (\textit{p}-values $<0.05$). The improvements in segmentation quality of the proposed approach over state-of-the-art methods were further supported by the visual comparisons on both datasets (Fig. \ref{fig:visual_comparison}).

\subsection{Supervised Contrastive Regularization Visualization}
\label{sec:supervised_contrastive_regularization_visualization}

\begin{figure}[t]
\centering
\begin{adjustbox}{width=\textwidth}
\begin{tikzpicture}

%% tSNE plots
    
\node[inner sep=0pt] at (0,2)
    {\includegraphics[width=.165\textwidth]{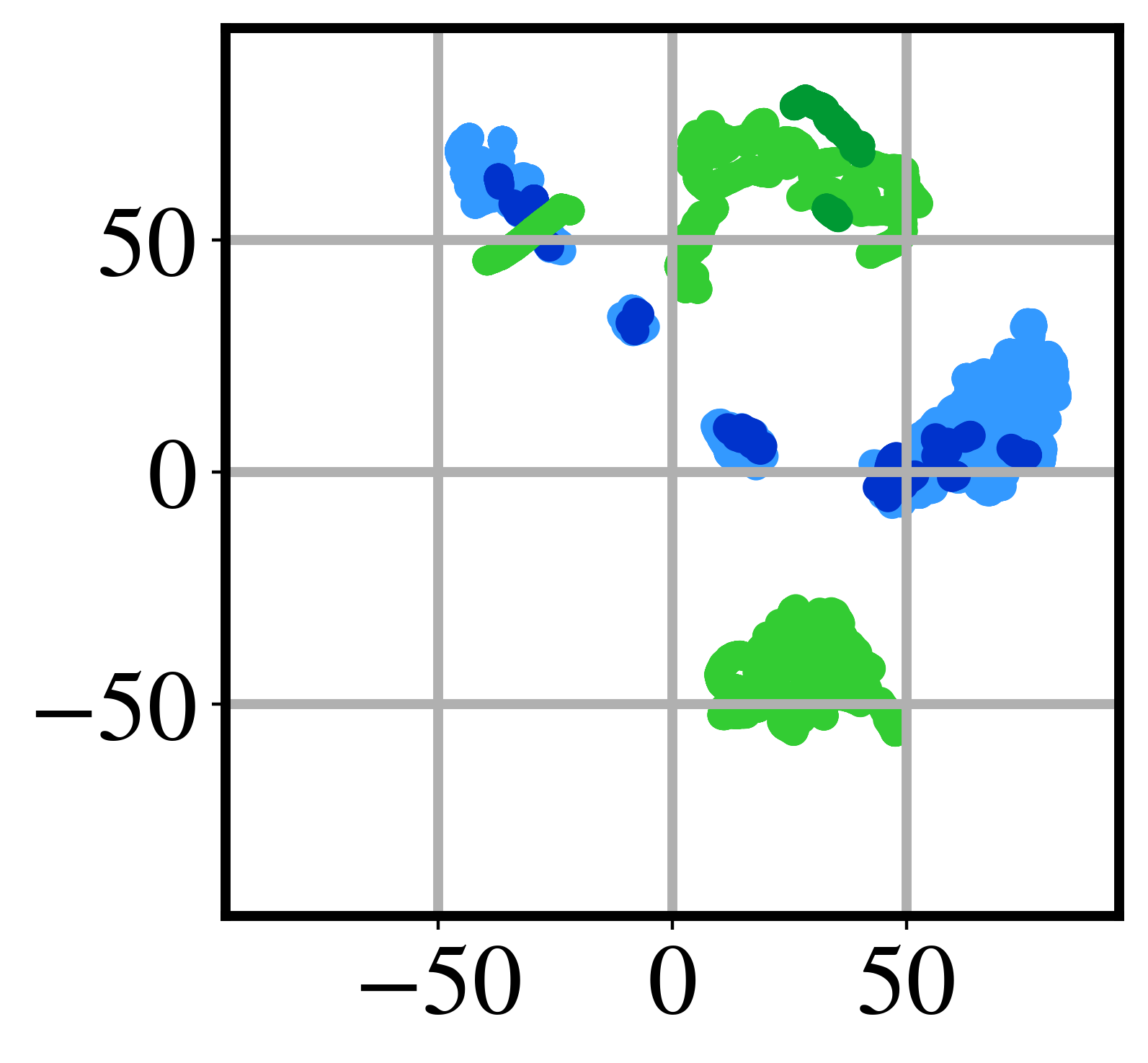}};
\node[inner sep=0pt] at (2,2)
    {\includegraphics[width=.165\textwidth]{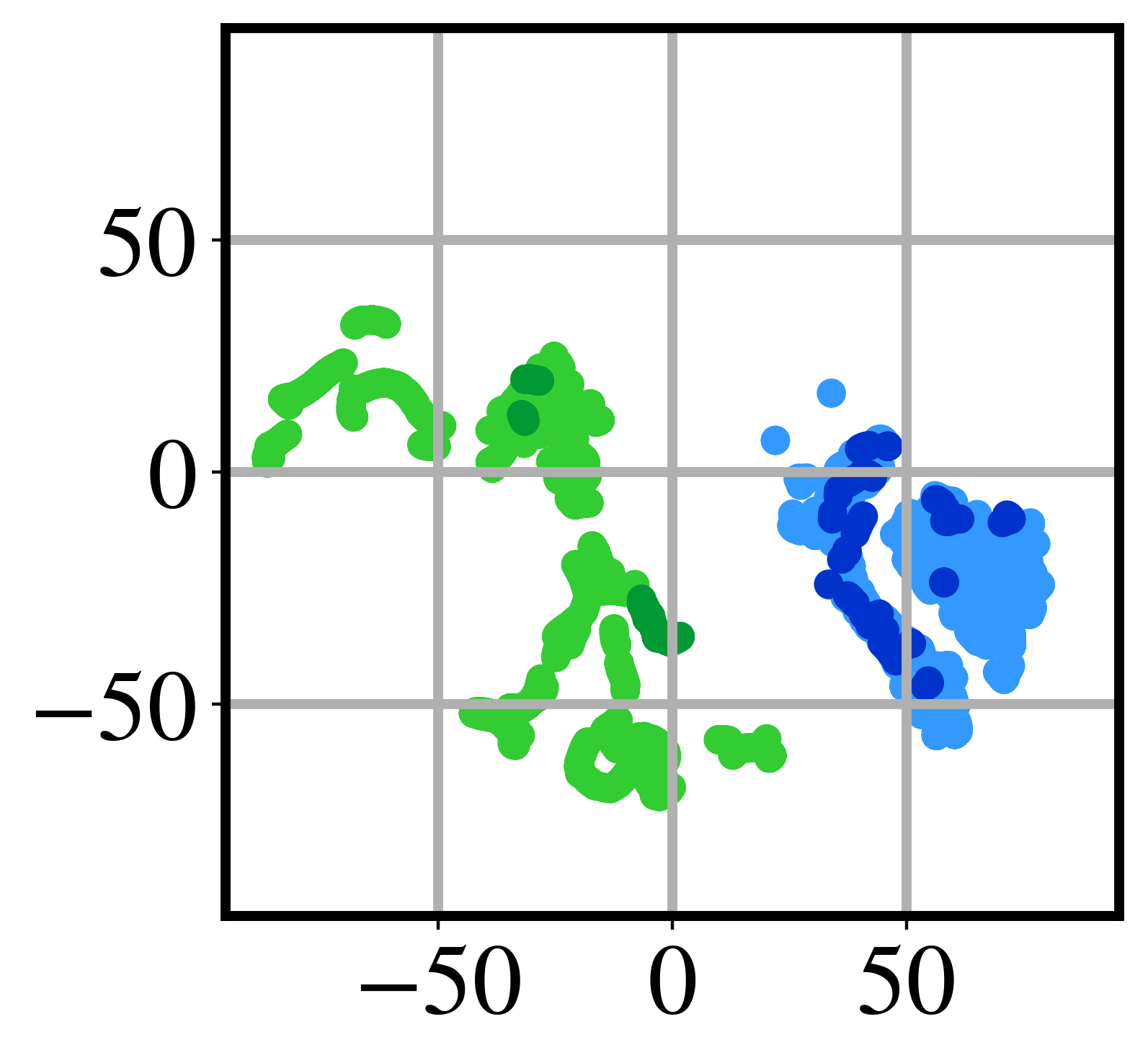}};
\node[inner sep=0pt] at (4,2)
    {\includegraphics[width=.165\textwidth]{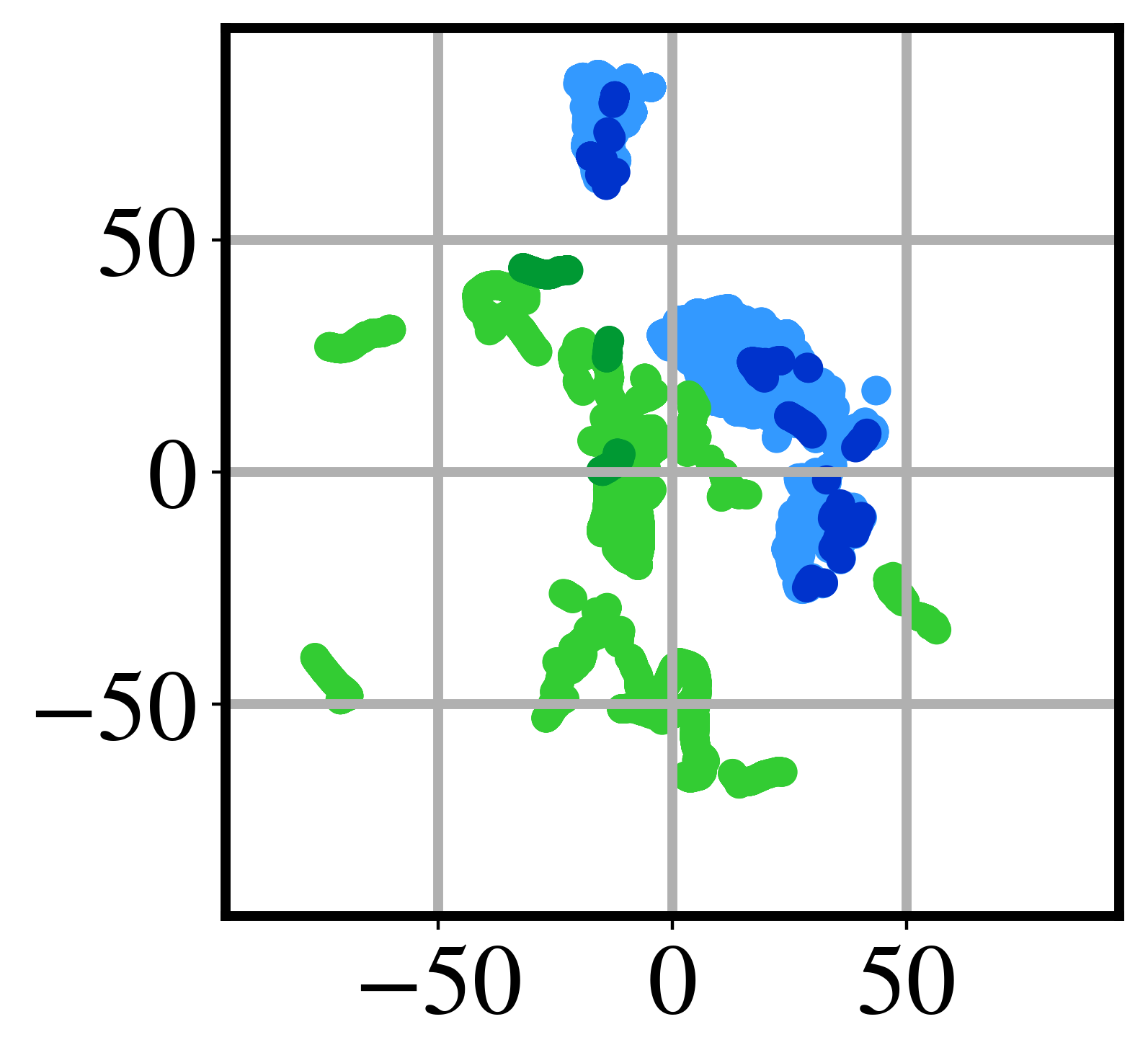}};
\node[inner sep=0pt] at (6,2)
    {\includegraphics[width=.165\textwidth]{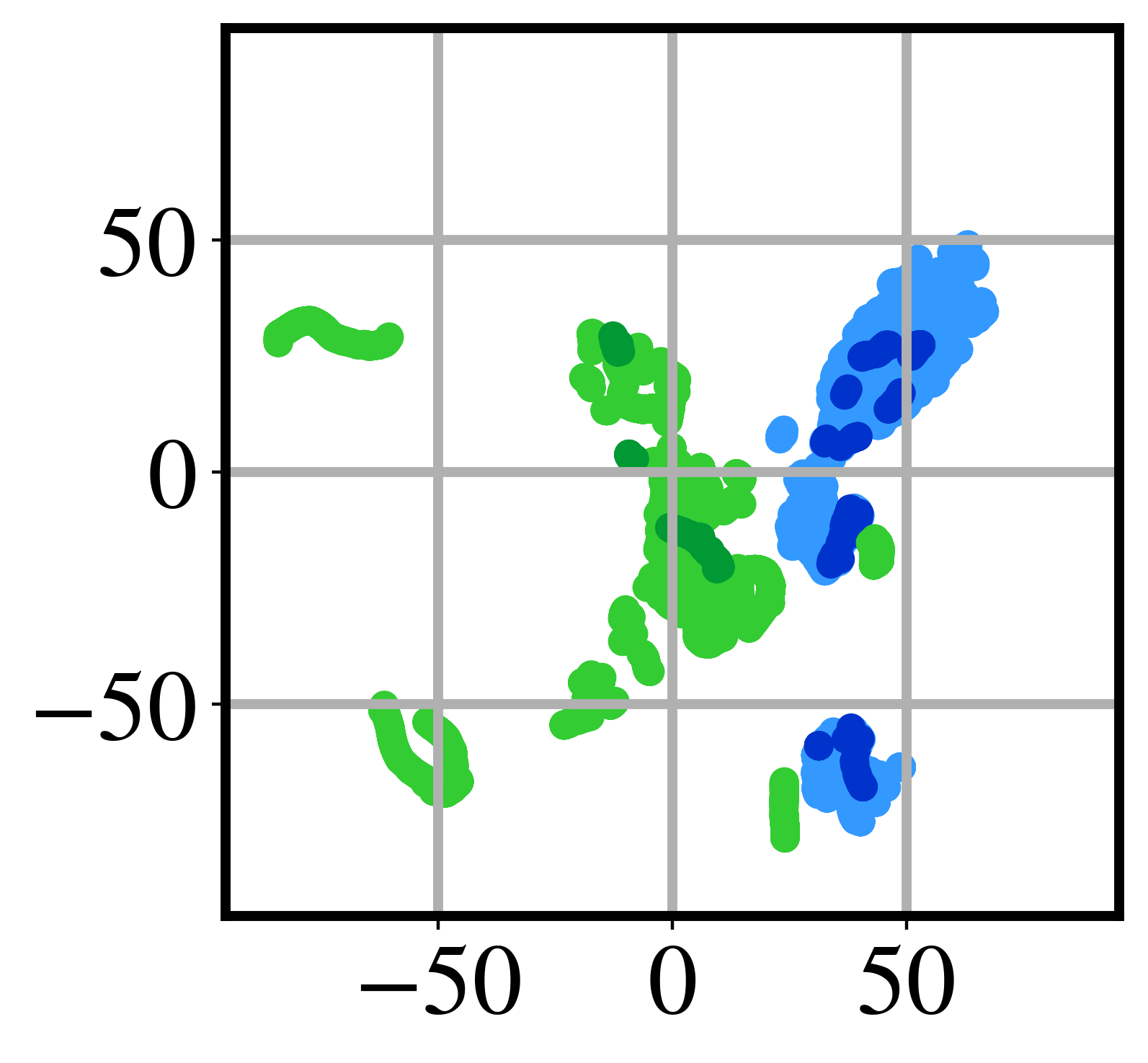}};
\node[inner sep=0pt] at (8,2)
    {\includegraphics[width=.165\textwidth]{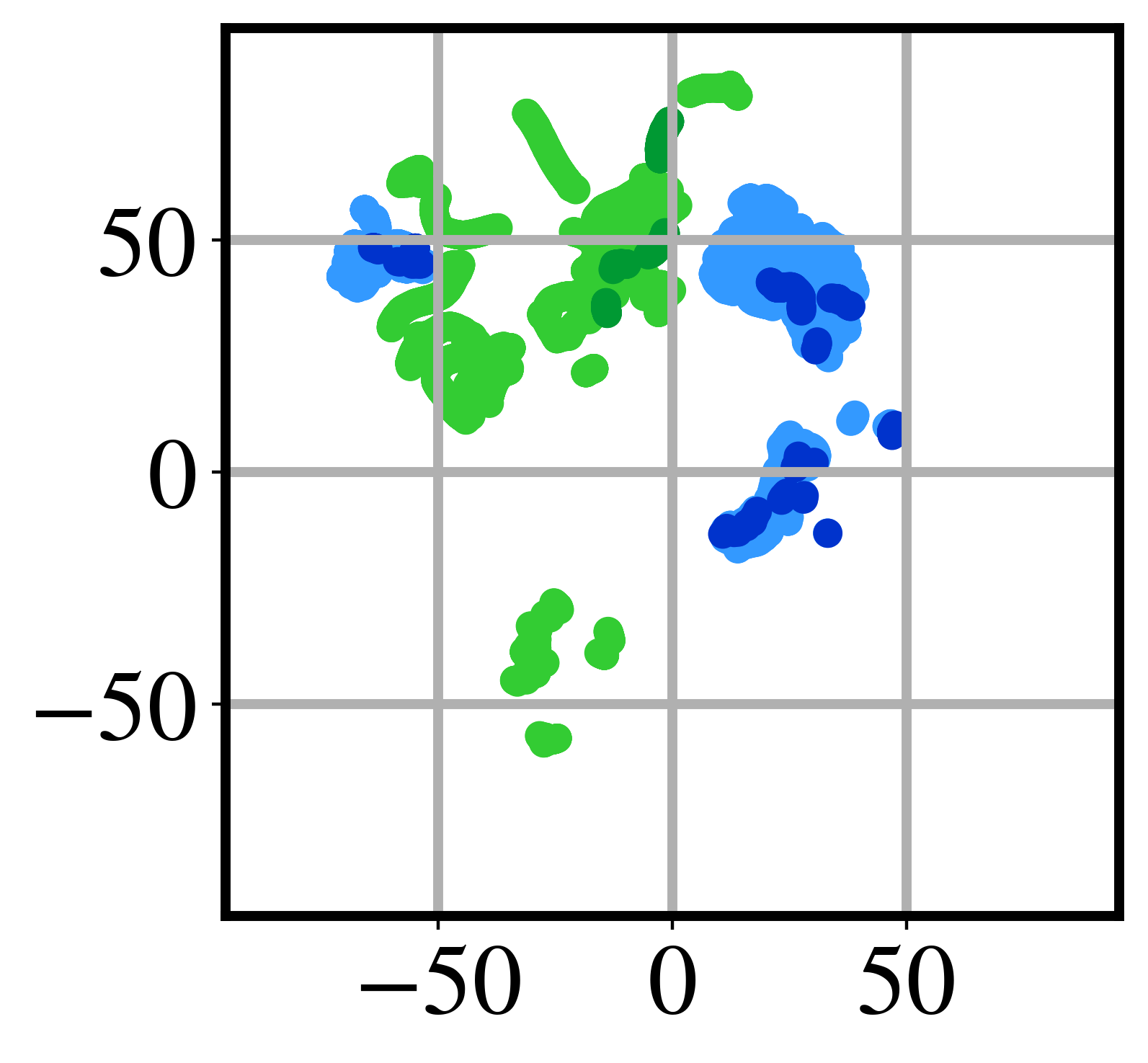}};
    
\node[inner sep=0pt] at (0,0)
    {\includegraphics[width=.165\textwidth]{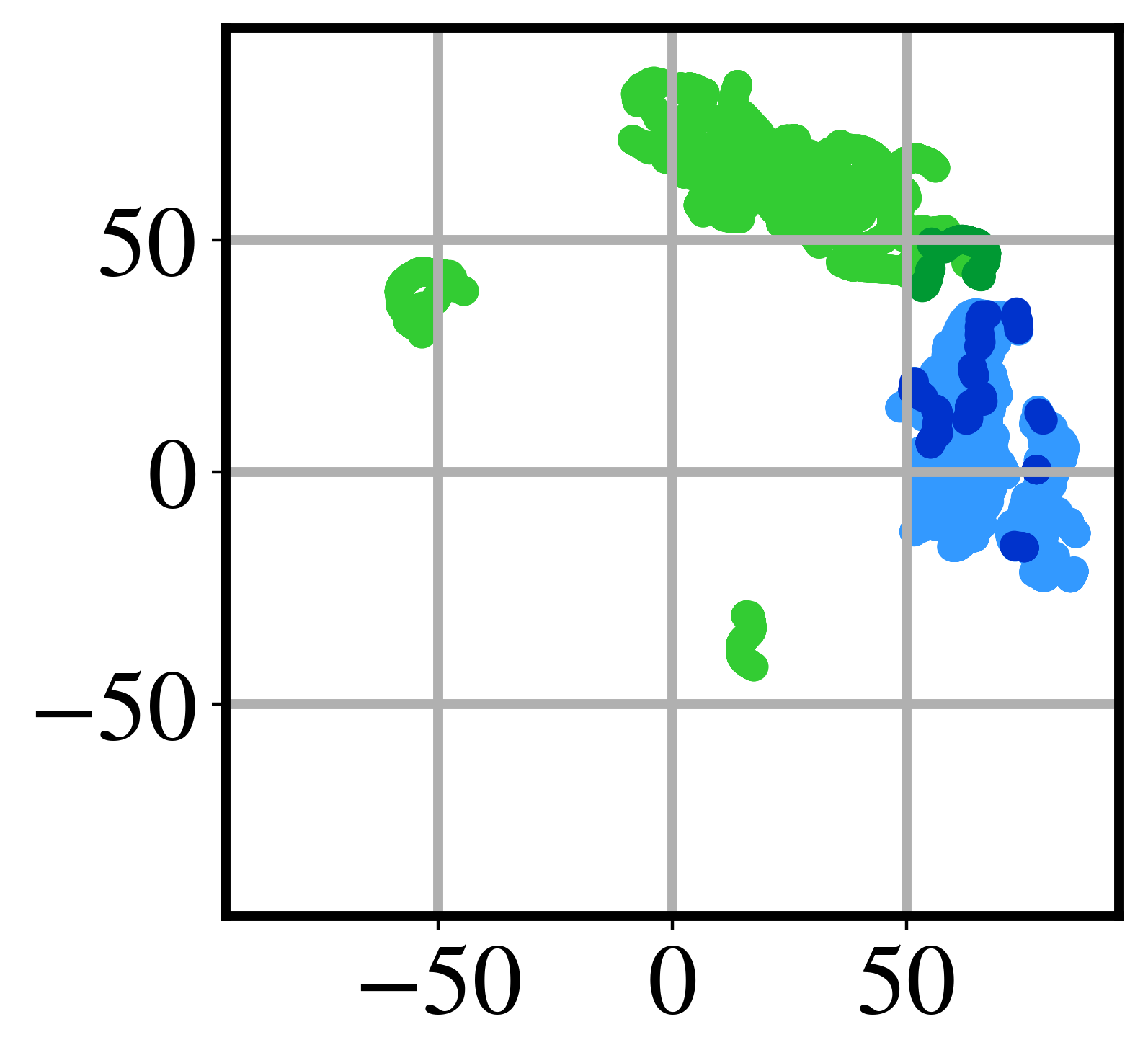}};
\node[inner sep=0pt] at (2,0)
    {\includegraphics[width=.165\textwidth]{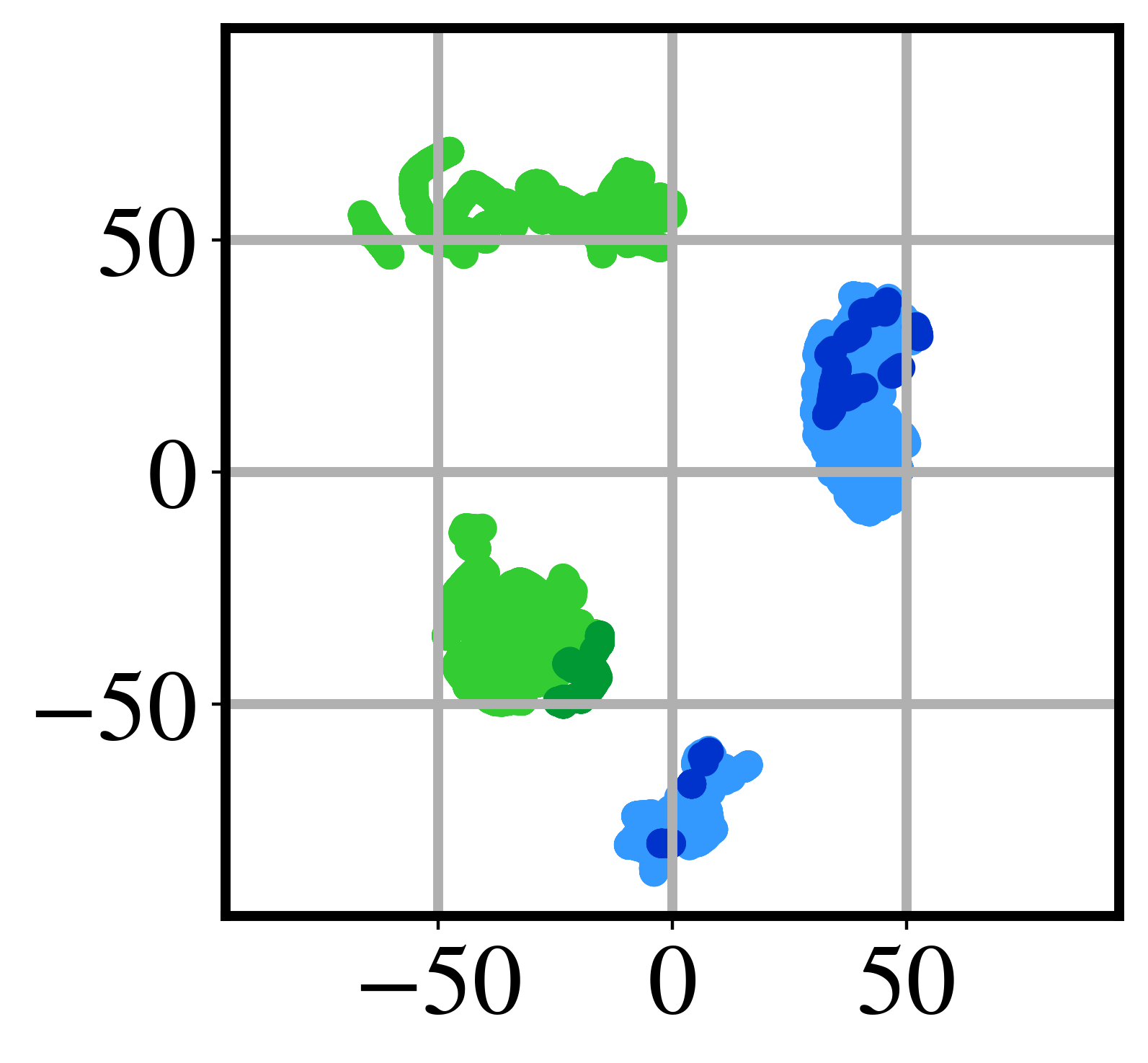}};
\node[inner sep=0pt] at (4,0)
    {\includegraphics[width=.165\textwidth]{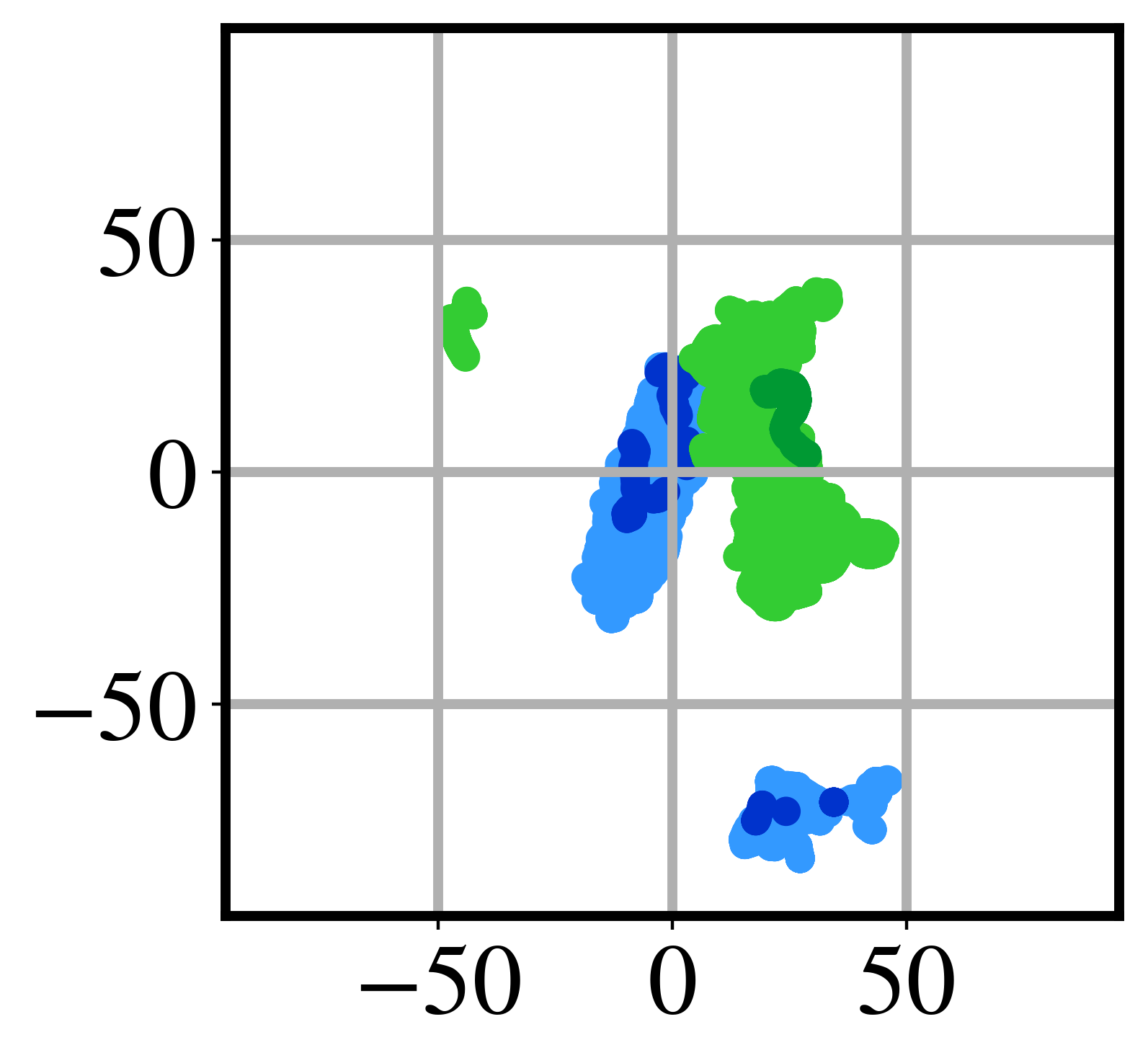}};
\node[inner sep=0pt] at (6,0)
    {\includegraphics[width=.165\textwidth]{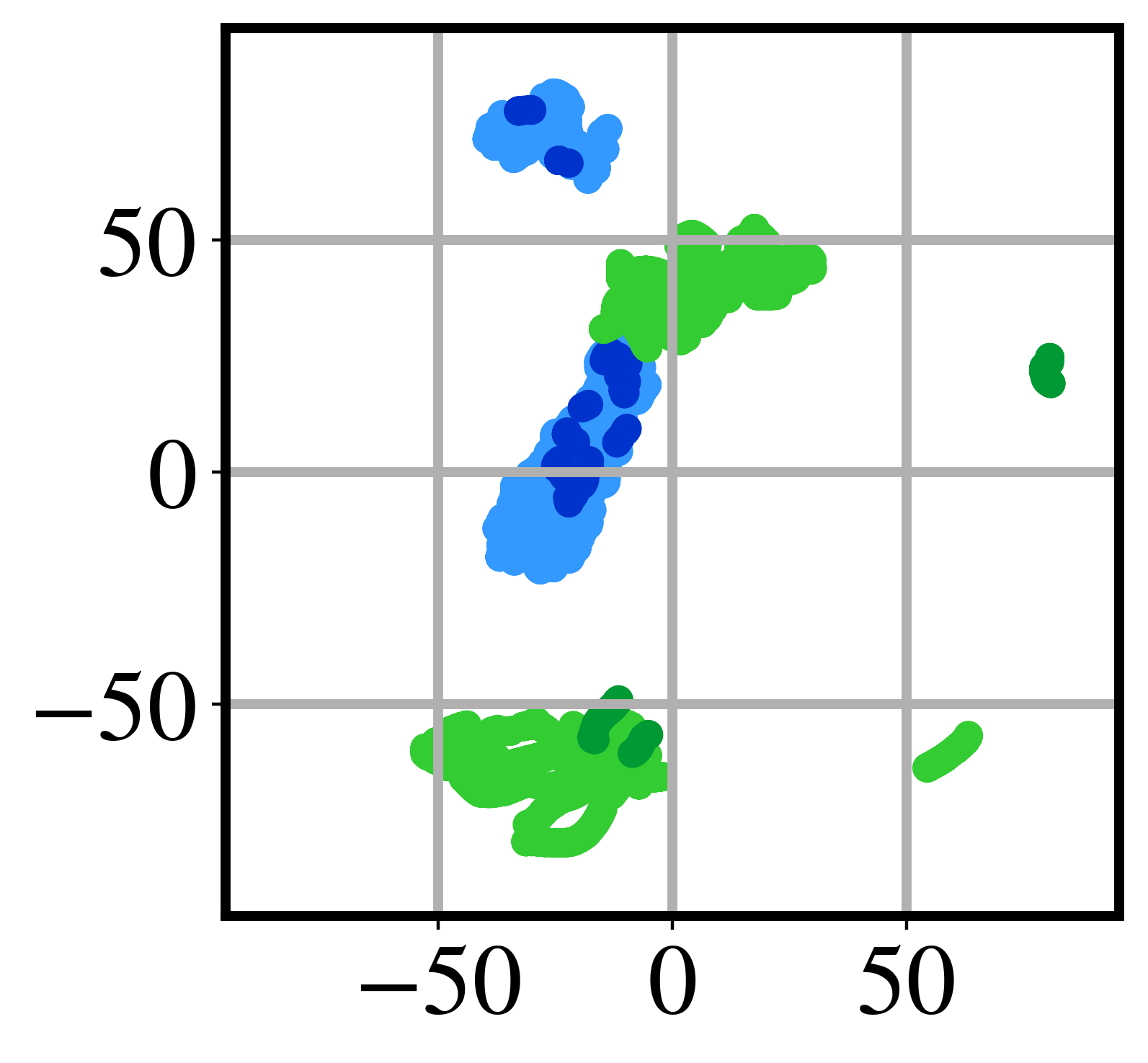}};
\node[inner sep=0pt] at (8,0)
    {\includegraphics[width=.165\textwidth]{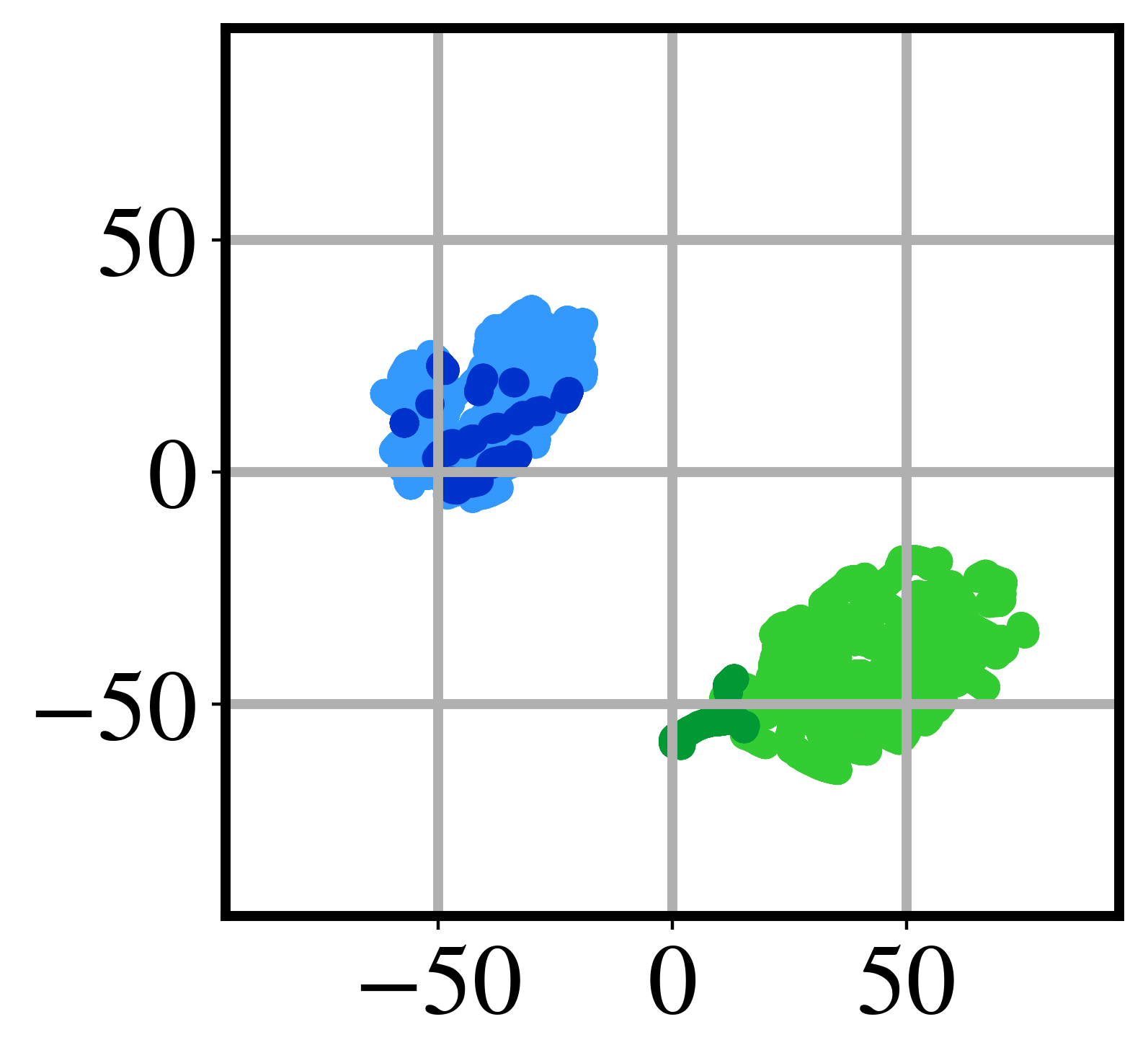}};

%% Legend
\node at (.175,3.05) {\scalebox{.85}{Epoch 10}};
\node at (2.175,3.05) {\scalebox{.85}{Epoch 15}};
\node at (4.175,3.05) {\scalebox{.85}{Epoch 20}};
\node at (6.175,3.05) {\scalebox{.85}{Epoch 25}};
\node at (8.175,3.05) {\scalebox{.85}{Epoch 30}};

\node[rotate=90] at (-1.1,2.075) {\scalebox{.85}{UNet DSL}};
\node[rotate=90] at (-1.4,.075) {\scalebox{.85}{UNet DSL}};
\node[rotate=90] at (-1.1,.075) {\scalebox{.85}{$+\mathcal{L}_{\text{Contrastive}}$}};

\node[inner sep=0pt] at (4.175,-1.35)
    {\includegraphics[width=.5\textwidth]{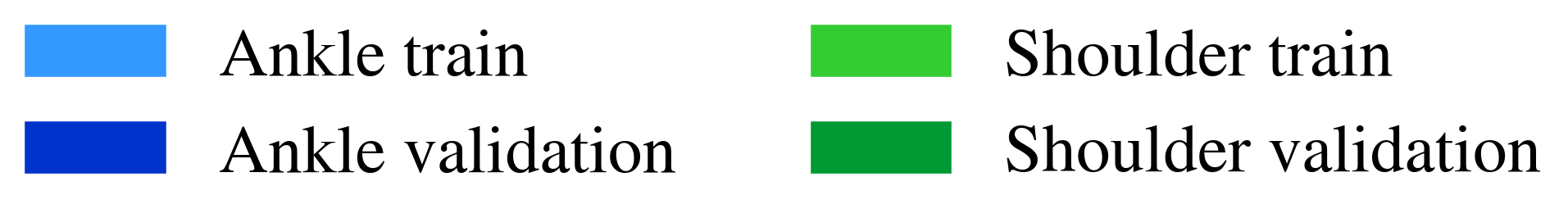}};

\draw[line width=.15mm] (1.15,-.95) rectangle (7.2,-1.7);

\end{tikzpicture}
\end{adjustbox}
  \caption{Visualization of the contrastive regularization which promotes intra-domain cohesion and inter-domain margins in embedded space during optimization. This visualization was obtained using the t-SNE algorithm in which each colored dot represented a 2D MR slice from the training or validation set of the ankle or shoulder datasets.}
  \label{fig:visualization_contrastive_regularization}
\end{figure}
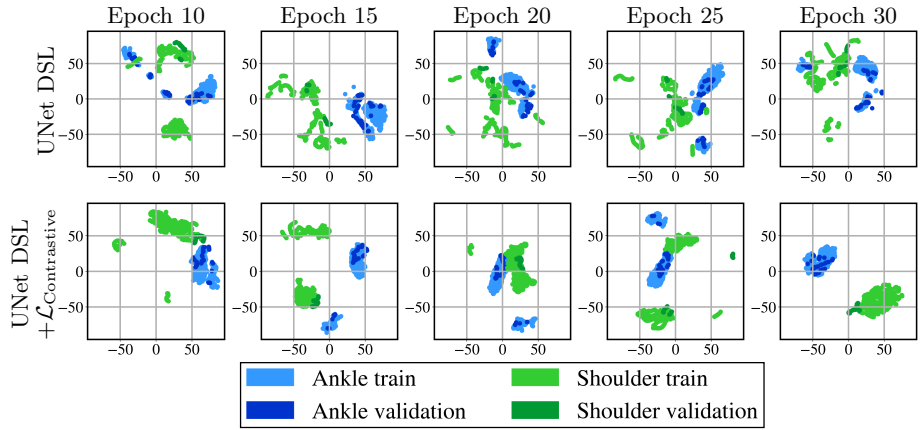

We performed an ablation study and compared the shared representation learnt by the proposed UNet DSL with and without contrastive regularization during optimization. The embedding of ankle and shoulder MR images from the training and validation set were visualized at epochs 10, 15, 20, 25 and 30. In order to visualize the 512 dimensional feature vectors, we applied a two-step dimensionality reduction as recommended in \cite{maaten_visualizing_2008}. We first employed principal component analysis which reduced the representations to 50 dimensional feature vectors, then the t-SNE algorithm embedded the data into a 2D space. The perplexity and learning rate of the t-SNE algorithm were set to 30 and 200 respectively.

Visualization of the shared representation provided a qualitative validation of the benefits of the additional supervised contrastive regularization on intra-domain cohesion and inter-domain separation. During optimization, the shared representation learnt by UNet DSL did not present margins between domains, while the addition of the contrastive regularization led to distinctive clusters. Hence, the shared representation of our proposed network was invariant to local variations and preserved the category of the domain. Moreover, the generalization capabilities of the network were visually attested as validation data-points were located inner their respective clusters (Fig. \ref{fig:visualization_contrastive_regularization}).

\section{Conclusion}
\label{sec:conclusion}

We presented a multi-task, multi-domain segmentation model with promising results for the task of pediatric bone segmentation in MR images. The proposed framework based on shared convolutional filters, domain-specific batch normalization and a domain-specific output segmentation layer, incorporated a contrastive regularization to enforce clustering in the shared representation and enhance generalization capabilities. An important perspective from this study is that this strategy can improve the segmentation performance on other musculoskeletal pediatric imaging datasets. Future studies aim at extending the segmentation model to other tissues such as muscles and cartilages, in order to provide a more complete description of the musculoskeletal system.

\subsubsection{Compliance with Ethical Standards.} MRI data acquisition on the pediatric cohort used in this study was performed in line with the principles of the Declaration of Helsinki. Ethical approval was granted by the Ethics Committee (Comit\'e Protection de Personnes Ouest VI) of CHRU Brest (2015-A01409-40).

\subsubsection{Acknowledgments.} This work was funded by IMT, Fondation Mines-Télécom and Institut Carnot TSN (Futur \& Ruptures). Data were acquired with the support of Fondation motrice (2015/7), Fondation de l’Avenir (AP-RM-16-041), PHRC 2015 (POPB 282) and Innoveo (CHRU Brest).

\bibliographystyle{splncs04}
\typeout{}
\bibliography{paper521}

\newpage 
\setcounter{table}{0}
\setcounter{section}{0}
\setcounter{figure}{0}
\setcounter{equation}{0}
\renewcommand{\thetable}{S\arabic{table}}
\renewcommand{\thesection}{S\arabic{section}}
\renewcommand{\thefigure}{S\arabic{figure}}
\renewcommand{\theequation}{S\arabic{equation}}

\section*{Supplementary Material}
\noindent \textbf{Metrics definitions.} Let $GT$ and $P$ be the ground truth and predicted 3D segmentation masks, and let $S_{GT}$ and $S_P$ be the surface voxels of the sets.
\begin{flalign*}
    &\textnormal{Dice} = \dfrac{2 \vert GT \cdot P\vert}{\vert GT \vert + \vert P \vert}, \enspace \textnormal{Sensitivity} = \dfrac{\vert GT \cdot P \vert}{\vert GT \vert}, \enspace \textnormal{Specificity} = \dfrac{\vert \overline{GT} \cdot \overline{P} \vert}{\vert \overline{GT} \vert}\\
    &\textnormal{ASSD} = \dfrac{1}{\vert S_{GT} \vert + \vert S_P \vert} \left(\sum_{s \in S_{GT}} \min_{s' \in S_P} \norm{s-s'}_2 + \sum_{s \in S_{P}} \min_{s' \in S_{GT}} \norm{s-s'}_2 \right)\\
    &\textnormal{MSSD} = \max (h(S_{GT}, S_P), h(S_P,S_{GT})), \enspace \textnormal{with} \; h(S,S') = \max_{s \in S} \min_{s' \in S'} \norm{s-s'}_2\\
    &\textnormal{RAVD} = \dfrac{\left| \vert GT \vert - \vert P \vert \right|}{\vert GT \vert}
\end{flalign*}

\begin{table}[ht!]
\caption{Quantitative assessment of UNet and Att-UNet using baseline, joint and DSL schemes, and the proposed UNet with DSL and contrastive regularization on ankle and shoulder datasets. Metrics include sensitivity ($\%$), specificity ($\%$) and relative absolute volume difference (mm). Best results are in bold.}
\centering
    \begin{tabular}{|P{.45cm}|P{1cm}||P{1.6cm}|P{1.6cm}|P{1.6cm}|P{1.6cm}|P{1.6cm}|P{1.6cm}|P{1.6cm}|P{1.6cm}|} 
    \hline
    \multicolumn{2}{|c||}{\multirow{2}{*}{Metric}} & \multicolumn{3}{c|}{Ankle} & \multicolumn{3}{c|}{Shoulder} \\\cline{3-8}
    \multicolumn{2}{|c||}{} & Sens $\uparrow$ & Spec $\uparrow$ & RAVD $\downarrow$ & Sens $\uparrow$  & Spec $\uparrow$ & RAVD $\downarrow$\\ 
    \hline\hline
     
    \multirow{3}{*}{\rotatebox[origin=c]{90}{UNet}} & Base & 91.6$\pm$5.0 & \textbf{99.9$\pm$0.1} & 8.9$\pm$5.0 & 79.8$\pm$16.5 & \textbf{99.9$\pm$0.1} & 18.9$\pm$23.7 \\\cline{2-8}
    & Joint & 92.1$\pm$5.0 & \textbf{99.9$\pm$0.1} & 7.7$\pm$5.1 & 80.5$\pm$14.9 & \textbf{99.9$\pm$0.1} & 13.7$\pm$15.6 \\\cline{2-8}
    & DSL & 92.2$\pm$4.2 & \textbf{99.9$\pm$0.1} & 7.3$\pm$3.8 & 81.4$\pm$15.2 & \textbf{99.9$\pm$0.1} & 15.4$\pm$15.2 \\\hline
    
    \multirow{3}{*}{\rotatebox[origin=c]{90}{Att}} & Base & 91.7$\pm$4.2 & \textbf{99.9$\pm$0.1} & 9.1$\pm$4.6 & 80.1$\pm$15.8 & \textbf{99.9$\pm$0.1} & 13.9$\pm$16.3 \\\cline{2-8}
    & Joint & 91.8$\pm$4.2 & \textbf{99.9$\pm$0.1} & 8.8$\pm$3.3 & 79.7$\pm$17.0 & \textbf{99.9$\pm$0.1} & 15.8$\pm$18.1 \\\cline{2-8}
    & DSL & \textbf{93.0$\pm$3.6} & \textbf{99.9$\pm$0.1} & 7.2$\pm$3.5 & 81.1$\pm$16.3 & \textbf{99.9$\pm$0.1} & 15.3$\pm$14.7 \\\hline\hline
  
    \multicolumn{2}{|c||}{Proposed} & 92.5$\pm$3.2 & \textbf{99.9$\pm$0.1} & \textbf{6.9$\pm$2.3} & \textbf{84.5$\pm$10.9} & \textbf{99.9$\pm$0.1} & \textbf{8.8$\pm$7.5}
    \\\hline
    
    \end{tabular}
\label{tab:results_supp}
\end{table}

\begin{table}[ht!]
\caption{Statistical analysis between the proposed model with UNet and Att-UNet using baseline, joint and DSL schemes, through Kolmogorov-Smirnov non-parametric test using sensitivity and specificity computed on 2D slices from ankle and shoulder datasets. Bold \textit{p}-values ($<0.05$) highlight statistically significant results.}
\centering
    \begin{tabular}{|P{1.25cm}||P{1.4cm}|P{1.4cm}|P{1.4cm}|P{1.4cm}|P{1.4cm}|P{1.4cm}|P{1.4cm}|} 
    \hline
    \multirow{2}{*}{Method} & \multicolumn{3}{c|}{UNet} & \multicolumn{3}{c|}{Att-UNet} & \multirow{2}{*}{Proposed} \\\cline{2-7}
    & Base & Joint & DSL & Base & Joint & DSL & \\\hline\hline
    
    Sens 2D & 88.0$\pm$18.3 & 88.3$\pm$18.3 & 89.1$\pm$17.2 & 88.1$\pm$18.3 & 88.0$\pm$18.2 & 89.1$\pm$16.8 & 90.0$\pm$16.7 \\\hline
    \textit{p}-value & \textbf{6.7\text{e-}21} & \textbf{2.3\text{e-}10} & \textbf{0.01} & \textbf{2.8\text{e-}16} & \textbf{1.5\text{e-}15} & \textbf{1.2\text{e-}6} & \--- \\\hline\hline
    
    Spec 2D & 99.9$\pm$0.1 & 99.9$\pm$0.1 & 99.9$\pm$0.1 & 99.9$\pm$0.1 & 99.9$\pm$0.1 & 99.9$\pm$0.1 & 99.9$\pm$0.1 \\\hline
    \textit{p}-value & \textbf{3.8\text{e-}5} & \textbf{1.1\text{e-}5} & \textbf{3.8\text{e-}5} & \textbf{8.4\text{e-}7} & \textbf{1.1\text{e-}5} & \textbf{0.01} & \--- \\\hline
    
    \end{tabular}
\label{tab:statistical_analsyis_supp}
\end{table}

\begin {figure*}[ht!]
\centering
\begin{adjustbox}{width=\textwidth}
\begin{tikzpicture}
\begin{scope}[spy using outlines=
      {circle, magnification=2.5, size=.36cm, connect spies, rounded corners}]

%% Pictures
%% Ankle 1
\node[inner sep=0pt] at (0,0)
    {\includegraphics[width=.115\textwidth]{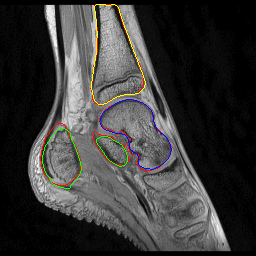}};
\node[inner sep=0pt] at (1.5,0)
    {\includegraphics[width=.115\textwidth]{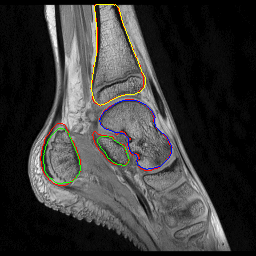}};
\node[inner sep=0pt] at (3,0)
    {\includegraphics[width=.115\textwidth]{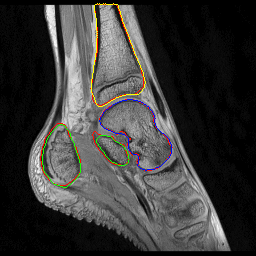}}; 
\node[inner sep=0pt] at (4.5,0)
    {\includegraphics[width=.115\textwidth]{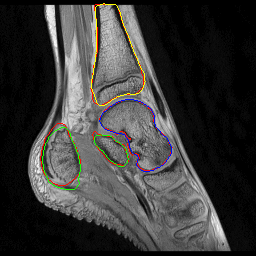}};
\node[inner sep=0pt] at (6,0)
    {\includegraphics[width=.115\textwidth]{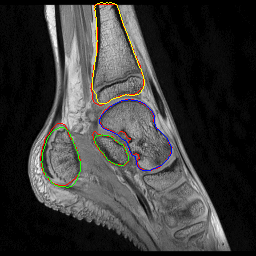}};
\node[inner sep=0pt] at (7.5,0)
    {\includegraphics[width=.115\textwidth]{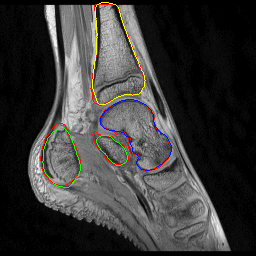}};
\node[inner sep=0pt] at (9,0)
    {\includegraphics[width=.115\textwidth]{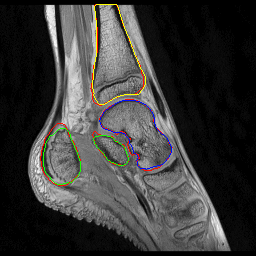}};
    
%% Ankle 2
\node[inner sep=0pt] at (0,-1.5)
    {\includegraphics[width=.115\textwidth]{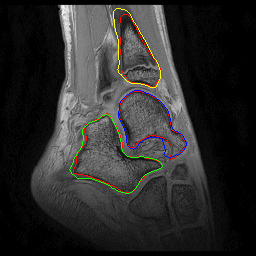}};
\node[inner sep=0pt] at (1.5,-1.5)
    {\includegraphics[width=.115\textwidth]{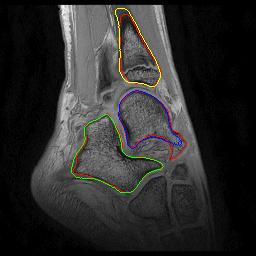}};
\node[inner sep=0pt] at (3,0-1.5)
    {\includegraphics[width=.115\textwidth]{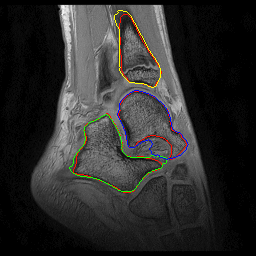}}; 
\node[inner sep=0pt] at (4.5,-1.5)
    {\includegraphics[width=.115\textwidth]{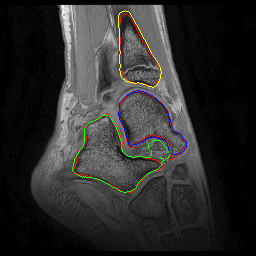}};
\node[inner sep=0pt] at (6,-1.5)
    {\includegraphics[width=.115\textwidth]{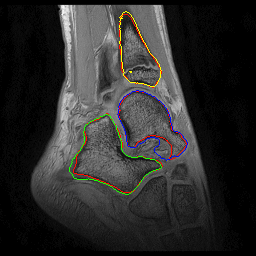}};
\node[inner sep=0pt] at (7.5,-1.5)
    {\includegraphics[width=.115\textwidth]{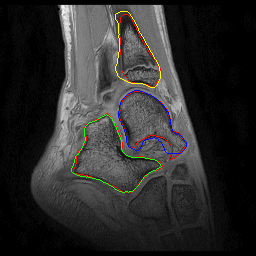}};
\node[inner sep=0pt] at (9,-1.5)
    {\includegraphics[width=.115\textwidth]{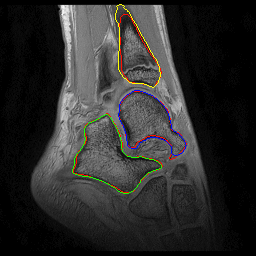}};
    
%% Shoulder 1
\node[inner sep=0pt] at (0,-3)
    {\includegraphics[width=.115\textwidth]{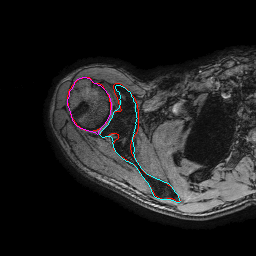}};
\node[inner sep=0pt] at (1.5,-3)
    {\includegraphics[width=.115\textwidth]{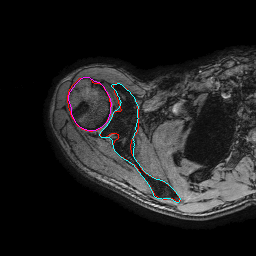}};
    \node[inner sep=0pt] at (3,-3)
{\includegraphics[width=.115\textwidth]{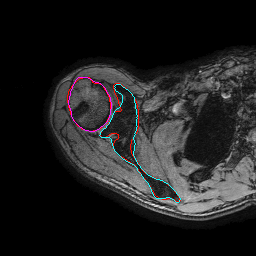}};
\node[inner sep=0pt] at (4.5,-3)
    {\includegraphics[width=.115\textwidth]{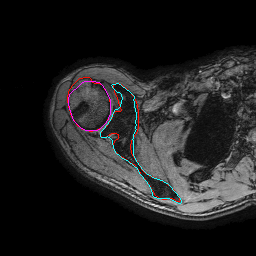}};
\node[inner sep=0pt] at (6,-3)
    {\includegraphics[width=.115\textwidth]{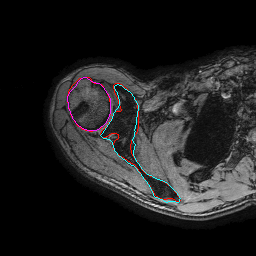}};
\node[inner sep=0pt] at (7.5,-3)
    {\includegraphics[width=.115\textwidth]{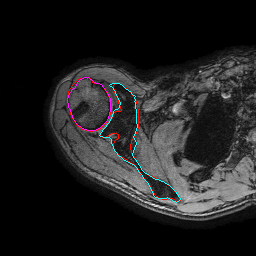}};
\node[inner sep=0pt] at (9,-3)
    {\includegraphics[width=.115\textwidth]{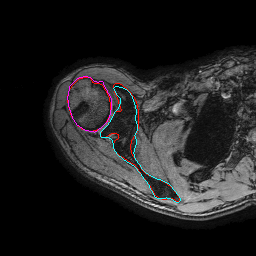}};
    
%% Shoulder 2
\node[inner sep=0pt] at (0,-4.5)
    {\includegraphics[width=.115\textwidth]{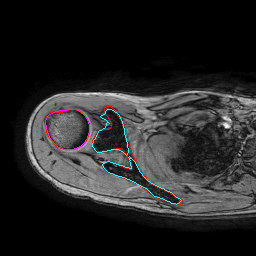}};
\node[inner sep=0pt] at (1.5,-4.5)
    {\includegraphics[width=.115\textwidth]{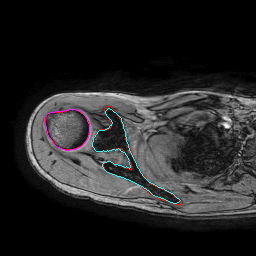}};
    \node[inner sep=0pt] at (3,-4.5)
{\includegraphics[width=.115\textwidth]{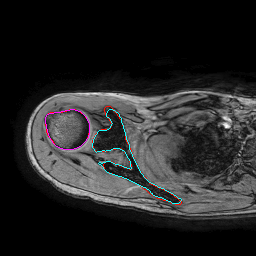}};
\node[inner sep=0pt] at (4.5,-4.5)
    {\includegraphics[width=.115\textwidth]{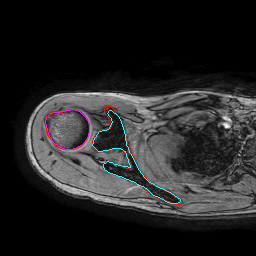}};
\node[inner sep=0pt] at (6,-4.5)
    {\includegraphics[width=.115\textwidth]{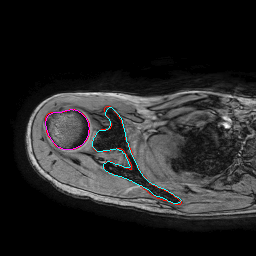}};
\node[inner sep=0pt] at (7.5,-4.5)
    {\includegraphics[width=.115\textwidth]{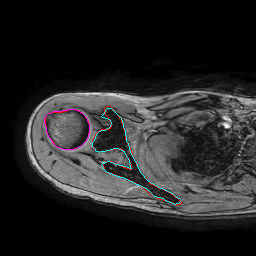}};
\node[inner sep=0pt] at (9,-4.5)
    {\includegraphics[width=.115\textwidth]{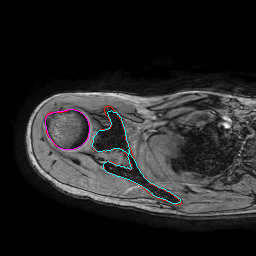}};
    
%% Zoom
%% Ankle 1-1
\spy [Dandelion] on (-.16,.16) in node [left] at (-.3,.48);
\spy [Dandelion] on (-.03,-.06) in node [left] at (.66,.48);
\spy [Dandelion] on (-.31,-.30) in node [left] at (.66,-.48);

%% Ankle 1-2
\spy [Dandelion] on (1.34,.16) in node [left] at (1.2,.48);
\spy [Dandelion] on (1.47,-.06) in node [left] at (2.16,.48);
\spy [Dandelion] on (1.19,-.30) in node [left] at (2.16,.-.48);

%% Ankle 1-3
\spy [Dandelion] on (2.84,.16) in node [left] at (2.7,.48);
\spy [Dandelion] on (2.97,-.06) in node [left] at (3.66,.48);
\spy [Dandelion] on (2.69,-.30) in node [left] at (3.66,.-.48);

%% Ankle 1-4
\spy [Dandelion] on (4.34,.16) in node [left] at (4.2,.48);
\spy [Dandelion] on (4.47,-.06) in node [left] at (5.16,.48);
\spy [Dandelion] on (4.19,-.30) in node [left] at (5.16,.-.48);

%% Ankle 1-5
\spy [Dandelion] on (5.84,.16) in node [left] at (5.7,.48);
\spy [Dandelion] on (5.97,-.06) in node [left] at (6.66,.48);
\spy [Dandelion] on (5.69,-.30) in node [left] at (6.66,.-.48);

%% Ankle 1-6
\spy [Dandelion] on (7.34,.16) in node [left] at (7.2,.48);
\spy [Dandelion] on (7.47,-.06) in node [left] at (8.16,.48);
\spy [Dandelion] on (7.19,-.30) in node [left] at (8.16,.-.48);

%% Ankle 1-7
\spy [Dandelion] on (8.84,.16) in node [left] at (8.7,.48);
\spy [Dandelion] on (8.97,-.06) in node [left] at (9.66,.48);
\spy [Dandelion] on (8.69,-.30) in node [left] at (9.66,.-.48);

%% Ankle 2-1
\spy [Dandelion] on (-.03,-.88) in node [left] at (-.3,-1.02);
\spy [Dandelion] on (.24,-1.63) in node [left] at (.66,-1.98);

%% Ankle 2-2
\spy [Dandelion] on (1.47,-.88) in node [left] at (1.2,-1.02);
\spy [Dandelion] on (1.74,-1.63) in node [left] at (2.16,.-1.98);

%% Ankle 2-3
\spy [Dandelion] on (2.97,-.88) in node [left] at (2.7,-1.02);
\spy [Dandelion] on (3.24,-1.63) in node [left] at (3.66,.-1.98);

%% Ankle 2-4
\spy [Dandelion] on (4.47,-.88) in node [left] at (4.2,-1.02);
\spy [Dandelion] on (4.74,-1.63) in node [left] at (5.16,.-1.98);

%% Ankle 2-5
\spy [Dandelion] on (5.97,-.88) in node [left] at (5.7,-1.02);
\spy [Dandelion] on (6.24,-1.63) in node [left] at (6.66,.-1.98);

%% Ankle 2-6
\spy [Dandelion] on (7.47,-.88) in node [left] at (7.2,-1.02);
\spy [Dandelion] on (7.74,-1.63) in node [left] at (8.16,.-1.98);

%% Ankle 2-7
\spy [Dandelion] on (8.97,-.88) in node [left] at (8.7,-1.02);
\spy [Dandelion] on (9.24,-1.63) in node [left] at (9.66,.-1.98);

%% Shoulder 1-1
\spy [Dandelion] on (-.27,-2.75) in node [left] at (-.3,-2.52);
\spy [Dandelion] on (-.07,-3.05) in node [left] at (-.3,-3.48);

%% Shoulder 1-2
\spy [Dandelion] on (1.23,-2.75) in node [left] at (1.2,-2.52);
\spy [Dandelion] on (1.43,-3.05) in node [left] at (1.2,-3.48);

%% Shoulder 1-3
\spy [Dandelion] on (2.73,-2.75) in node [left] at (2.7,-2.52);
\spy [Dandelion] on (2.93,-3.05) in node [left] at (2.7,-3.48);

%% Shoulder 1-4
\spy [Dandelion] on (4.23,-2.75) in node [left] at (4.2,-2.52);
\spy [Dandelion] on (4.43,-3.05) in node [left] at (4.2,-3.48);

%% Shoulder 1-5
\spy [Dandelion] on (5.73,-2.75) in node [left] at (5.7,-2.52);
\spy [Dandelion] on (5.93,-3.05) in node [left] at (5.7,-3.48);

%% Shoulder 1-6
\spy [Dandelion] on (7.23,-2.75) in node [left] at (7.2,-2.52);
\spy [Dandelion] on (7.43,-3.05) in node [left] at (7.2,-3.48);

%% Shoulder 1-7
\spy [Dandelion] on (8.73,-2.75) in node [left] at (8.7,-2.52);
\spy [Dandelion] on (8.93,-3.05) in node [left] at (8.7,-3.48);

%% Shoulder 2-1
\spy [Dandelion] on (-.43,-4.44) in node [left] at (-.3,-4.02);
\spy [Dandelion] on (-.1,-4.43) in node [left] at (.66,-4.02);
\spy [Dandelion] on (.01,-4.65) in node [left] at (.66,-4.98);

%% Shoulder 2-2
\spy [Dandelion] on (1.07,-4.44) in node [left] at (1.2,-4.02);
\spy [Dandelion] on (1.4,-4.43) in node [left] at (2.16,-4.02);
\spy [Dandelion] on (1.51,-4.65) in node [left] at (2.16,-4.98);

%% Shoulder 2-3
\spy [Dandelion] on (2.57,-4.44) in node [left] at (2.7,-4.02);
\spy [Dandelion] on (2.9,-4.43) in node [left] at (3.66,-4.02);
\spy [Dandelion] on (3.01,-4.65) in node [left] at (3.66,-4.98);

%% Shoulder 2-4
\spy [Dandelion] on (4.07,-4.44) in node [left] at (4.2,-4.02);
\spy [Dandelion] on (4.4,-4.43) in node [left] at (5.16,-4.02);
\spy [Dandelion] on (4.51,-4.65) in node [left] at (5.16,-4.98);

%% Shoulder 2-5
\spy [Dandelion] on (5.57,-4.44) in node [left] at (5.7,-4.02);
\spy [Dandelion] on (5.9,-4.43) in node [left] at (6.66,-4.02);
\spy [Dandelion] on (6.01,-4.65) in node [left] at (6.66,-4.98);

%% Shoulder 2-6
\spy [Dandelion] on (7.07,-4.44) in node [left] at (7.2,-4.02);
\spy [Dandelion] on (7.4,-4.43) in node [left] at (8.16,-4.02);
\spy [Dandelion] on (7.51,-4.65) in node [left] at (8.16,-4.98);

%% Shoulder 2-7
\spy [Dandelion] on (8.57,-4.44) in node [left] at (8.7,-4.02);
\spy [Dandelion] on (8.9,-4.43) in node [left] at (9.66,-4.02);
\spy [Dandelion] on (9.01,-4.65) in node [left] at (9.66,-4.98);

\end{scope}
    
%% Legend
\node at (0, 1.05) {\scalebox{.7}{UNet}};
\node at (0, .825) {\scalebox{.7}{Base}};
\node at (1.5, 1.05) {\scalebox{.7}{UNet}};
\node at (1.5, .825) {\scalebox{.7}{Joint}};
\node at (3, 1.05) {\scalebox{.7}{UNet}};
\node at (3, .825) {\scalebox{.7}{DSL}};
\node at (4.5, 1.05) {\scalebox{.7}{Att-UNet}};
\node at (4.5, .825) {\scalebox{.7}{Base}};
\node at (6, 1.05) {\scalebox{.7}{Att-UNet}};
\node at (6, .825) {\scalebox{.7}{Joint}};
\node at (7.5, 1.05) {\scalebox{.7}{Att-UNet}};
\node at (7.5, .825) {\scalebox{.7}{DSL}};
\node at (9, .825) {\scalebox{.7}{Proposed}};

\end{tikzpicture}
\end{adjustbox}
\caption{Visual comparison of UNet and Att-UNet using baseline, joint and DSL schemes, and the proposed model on ankle and shoulder datasets. Ground truth delineations are in red (\textcolor{red}{\---}). Predicted calcaneus, talus, tibia, humerus and scapula bones respectively appear in green (\textcolor{green}{\---}), blue (\textcolor{blue}{\---}), yellow (\textcolor{yellow}{\---}), magenta (\textcolor{magenta}{\---}) and cyan (\textcolor{cyan}{\---}).}
\label{fig:visual_comparison_supp}
\end{figure*}

\begin{figure}[ht!]
\centering
\begin{adjustbox}{width=\textwidth}
\begin{tikzpicture}

%% tSNE plots
\node[inner sep=0pt] at (0,0)
    {\includegraphics[width=.165\textwidth]{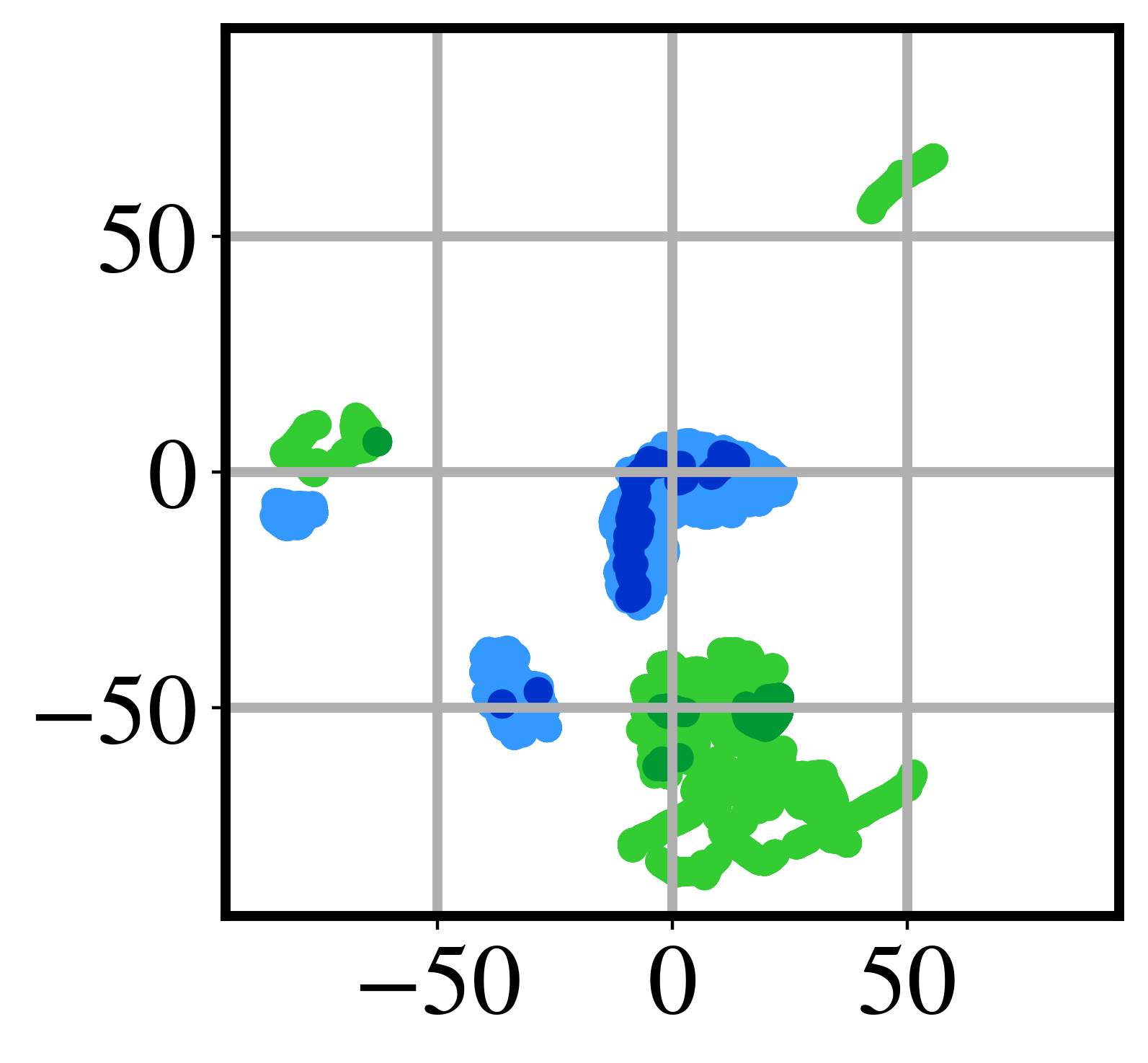}};
\node[inner sep=0pt] at (2,0)
    {\includegraphics[width=.165\textwidth]{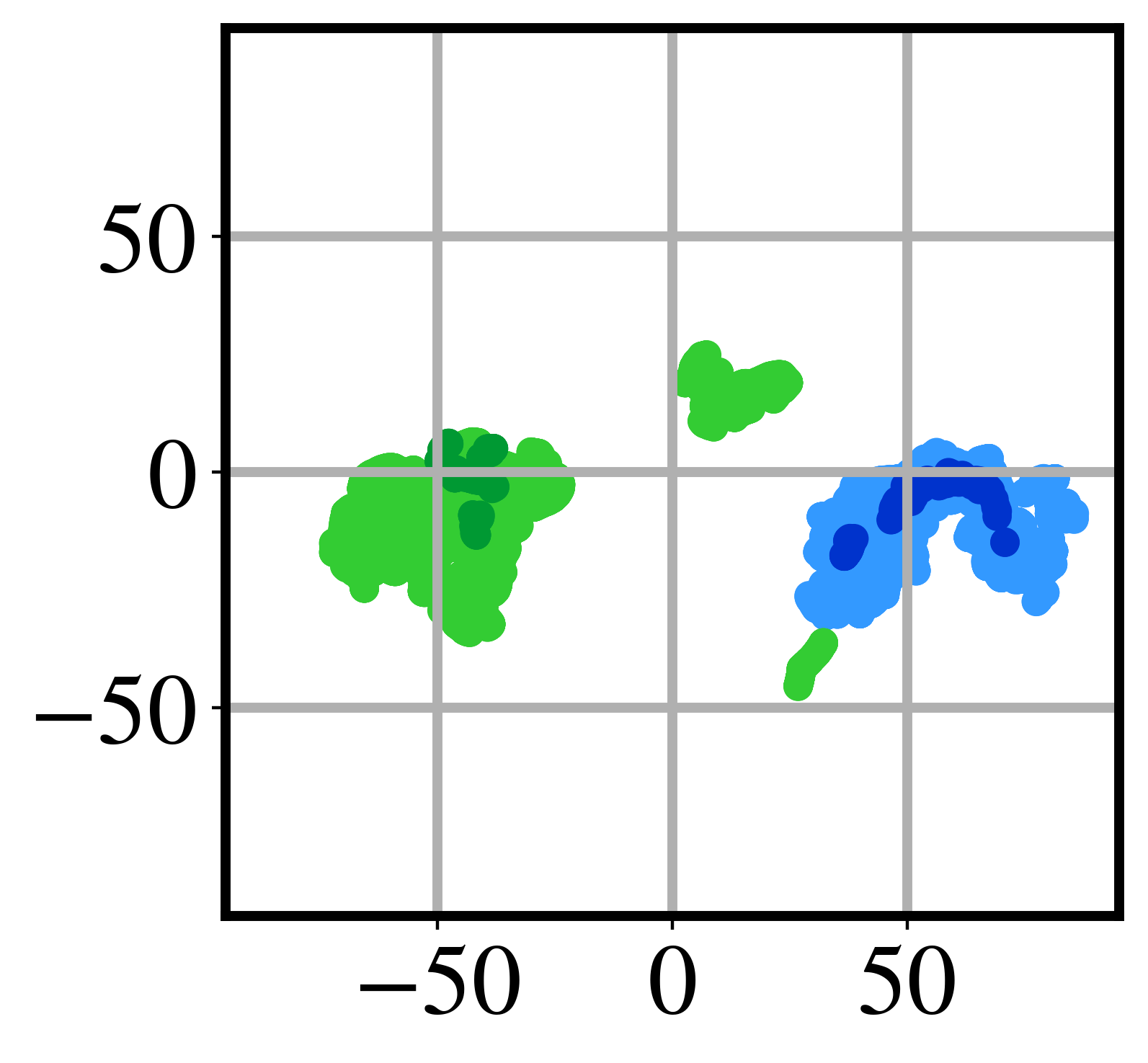}};
\node[inner sep=0pt] at (4,0)
    {\includegraphics[width=.165\textwidth]{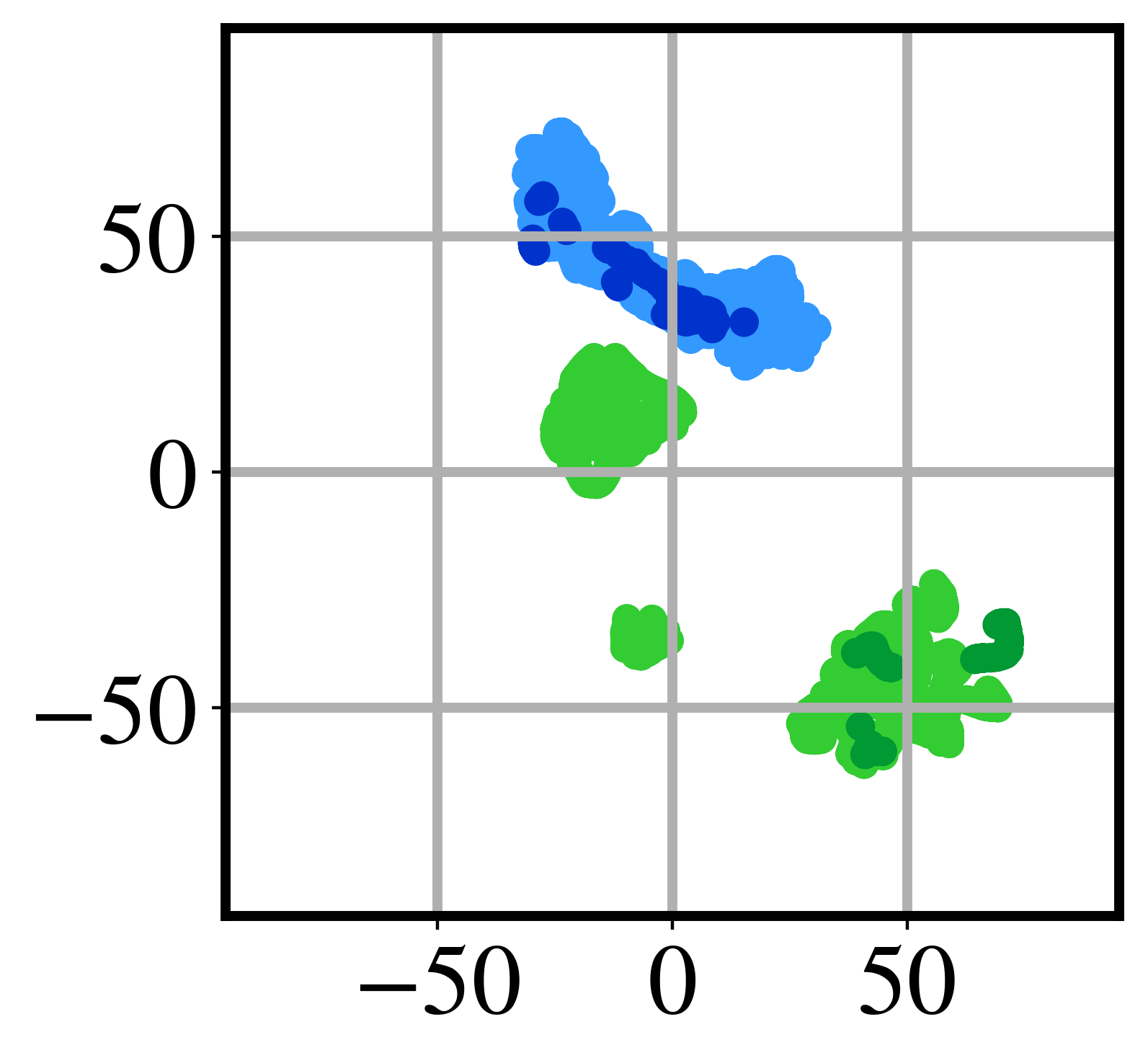}};
\node[inner sep=0pt] at (6,0)
    {\includegraphics[width=.165\textwidth]{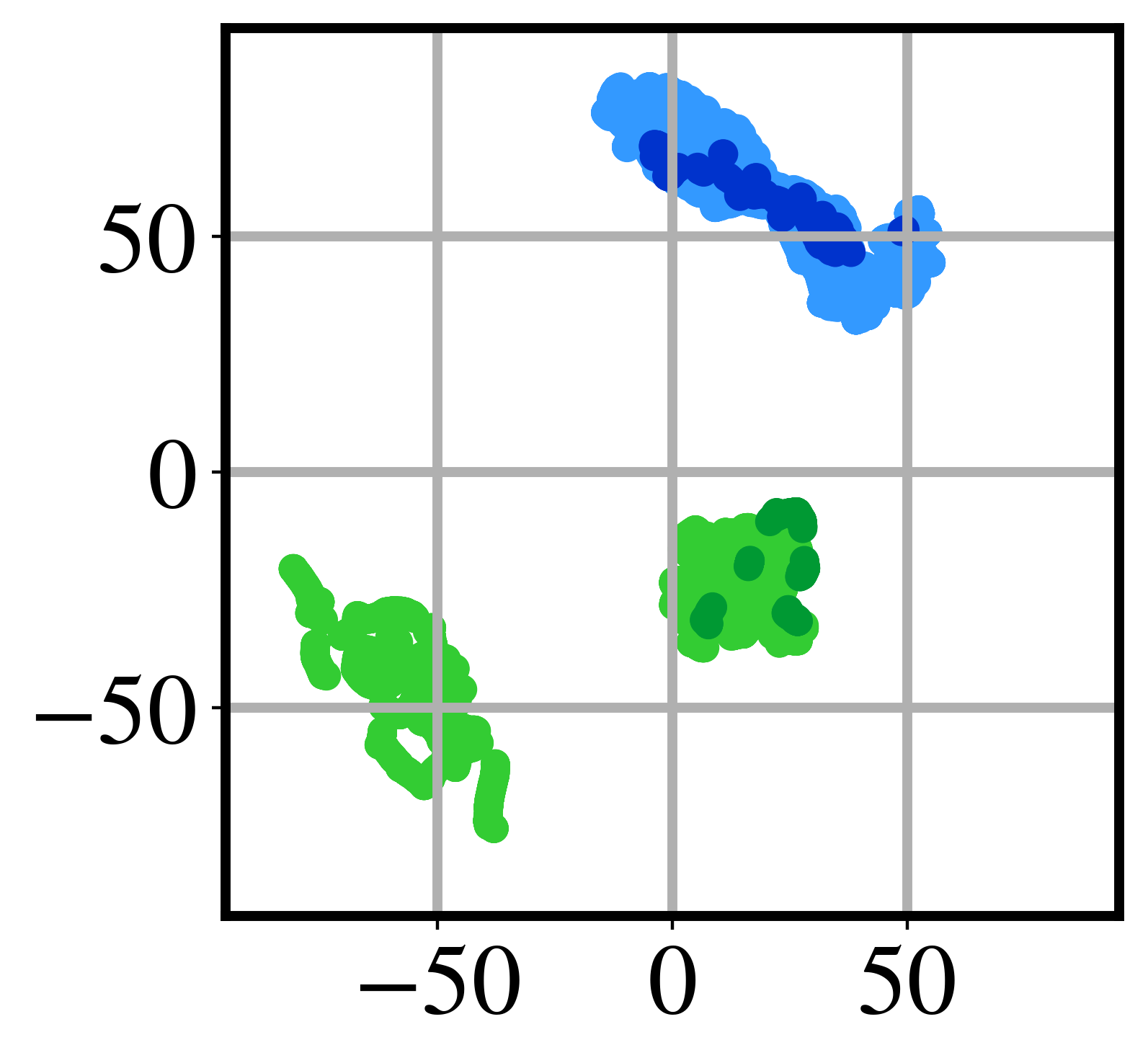}};
\node[inner sep=0pt] at (8,0)
    {\includegraphics[width=.165\textwidth]{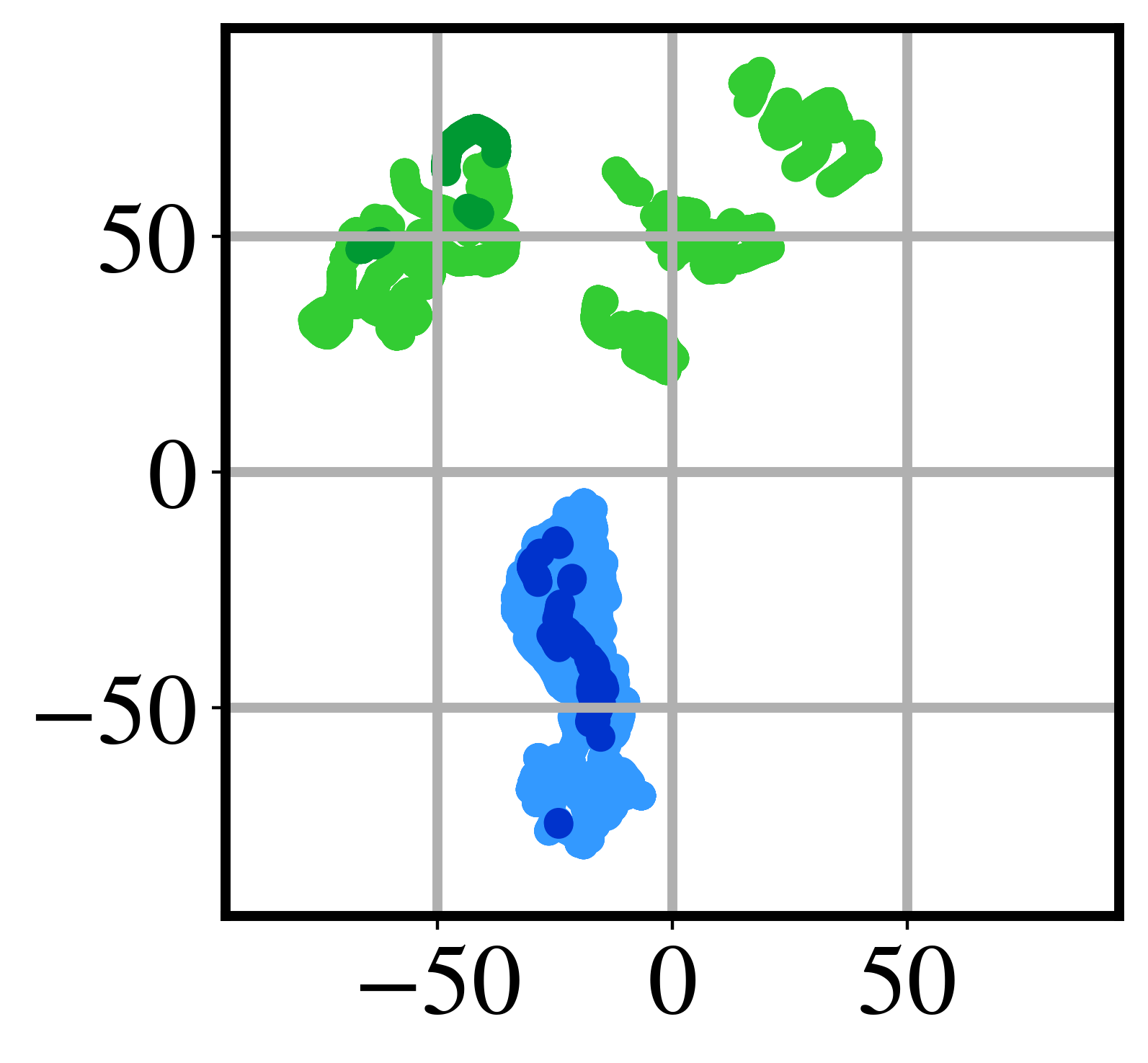}};
    
\node[inner sep=0pt] at (0,2)
    {\includegraphics[width=.165\textwidth]{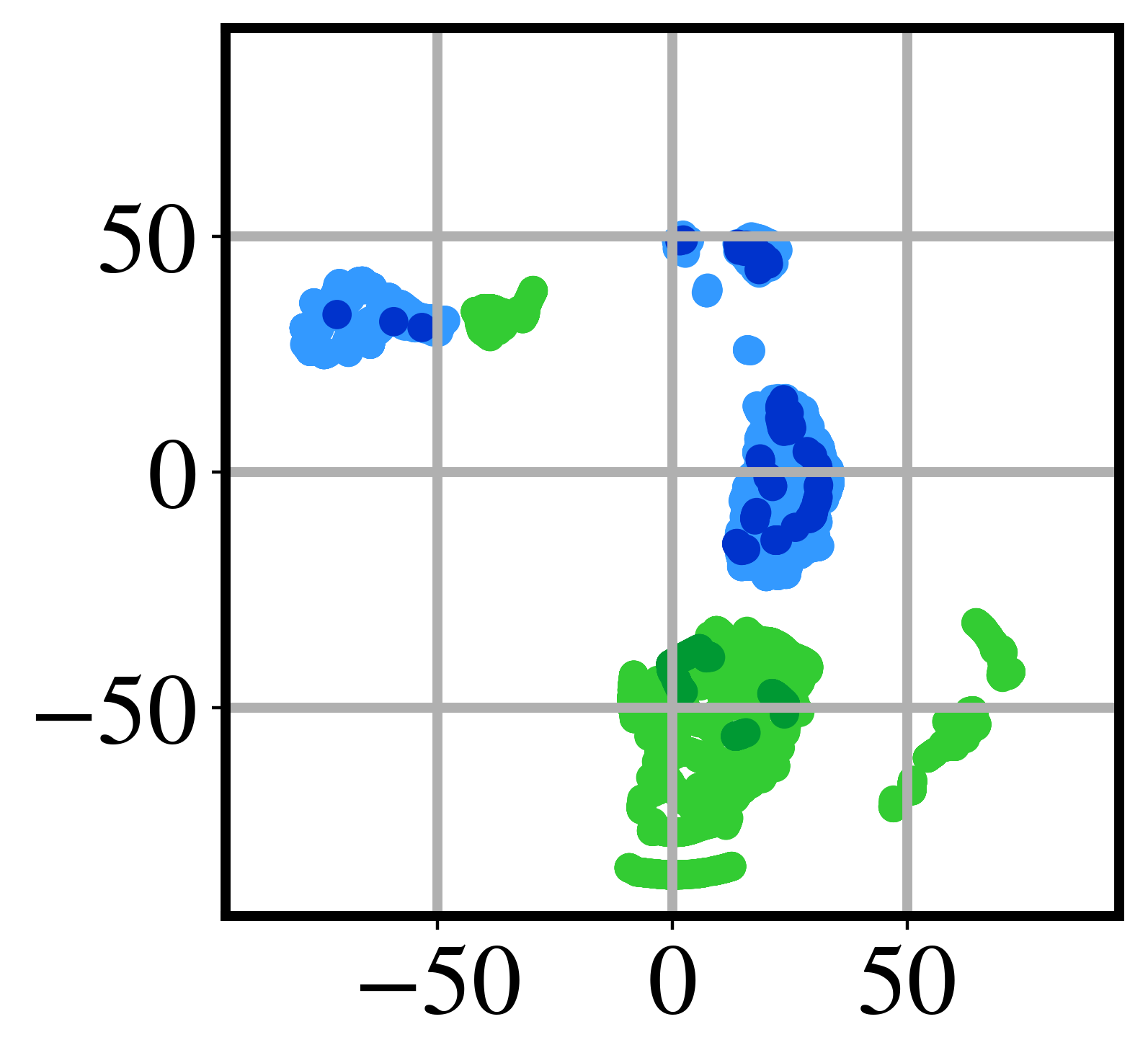}};
\node[inner sep=0pt] at (2,2)
    {\includegraphics[width=.165\textwidth]{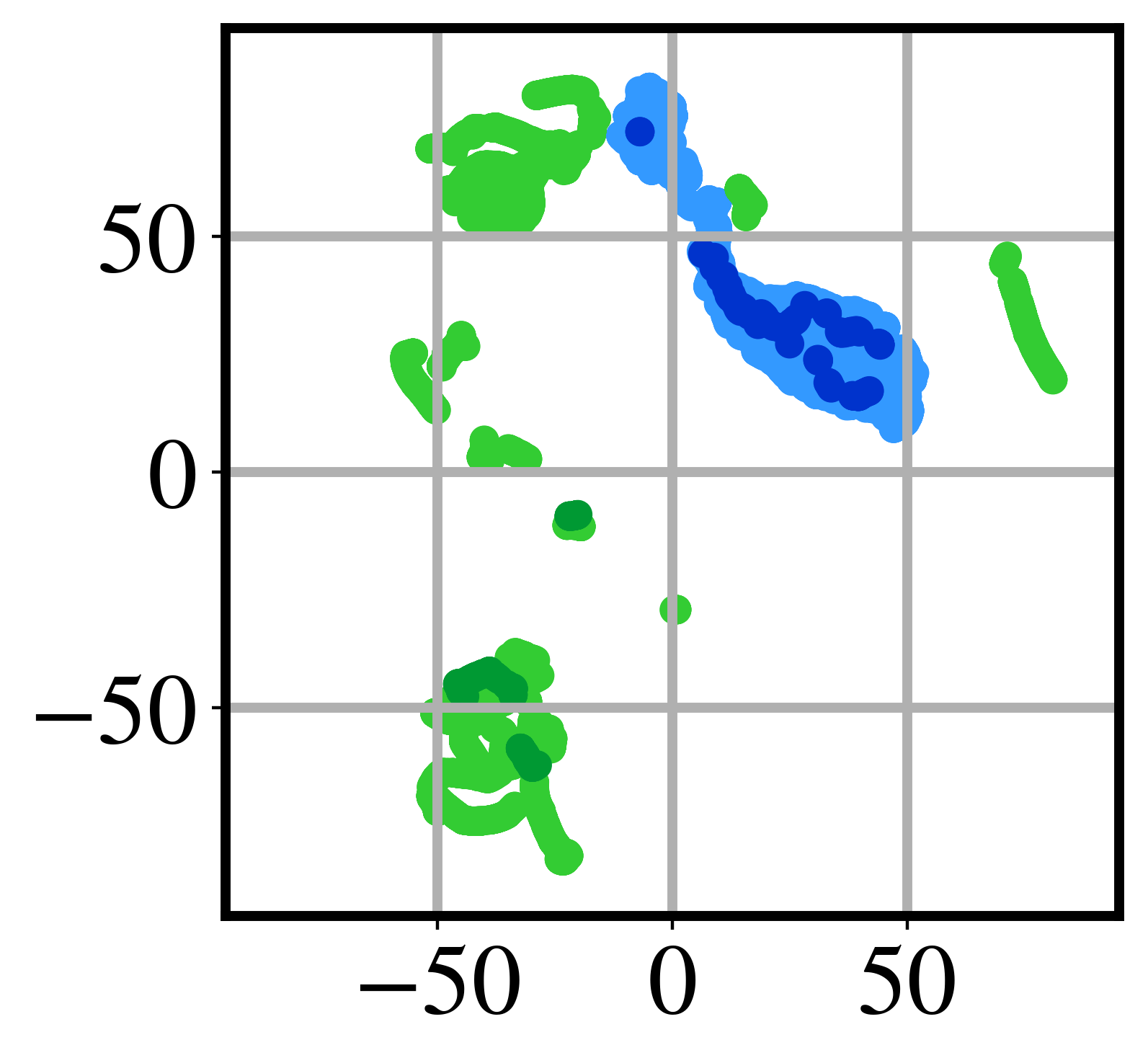}};
\node[inner sep=0pt] at (4,2)
    {\includegraphics[width=.165\textwidth]{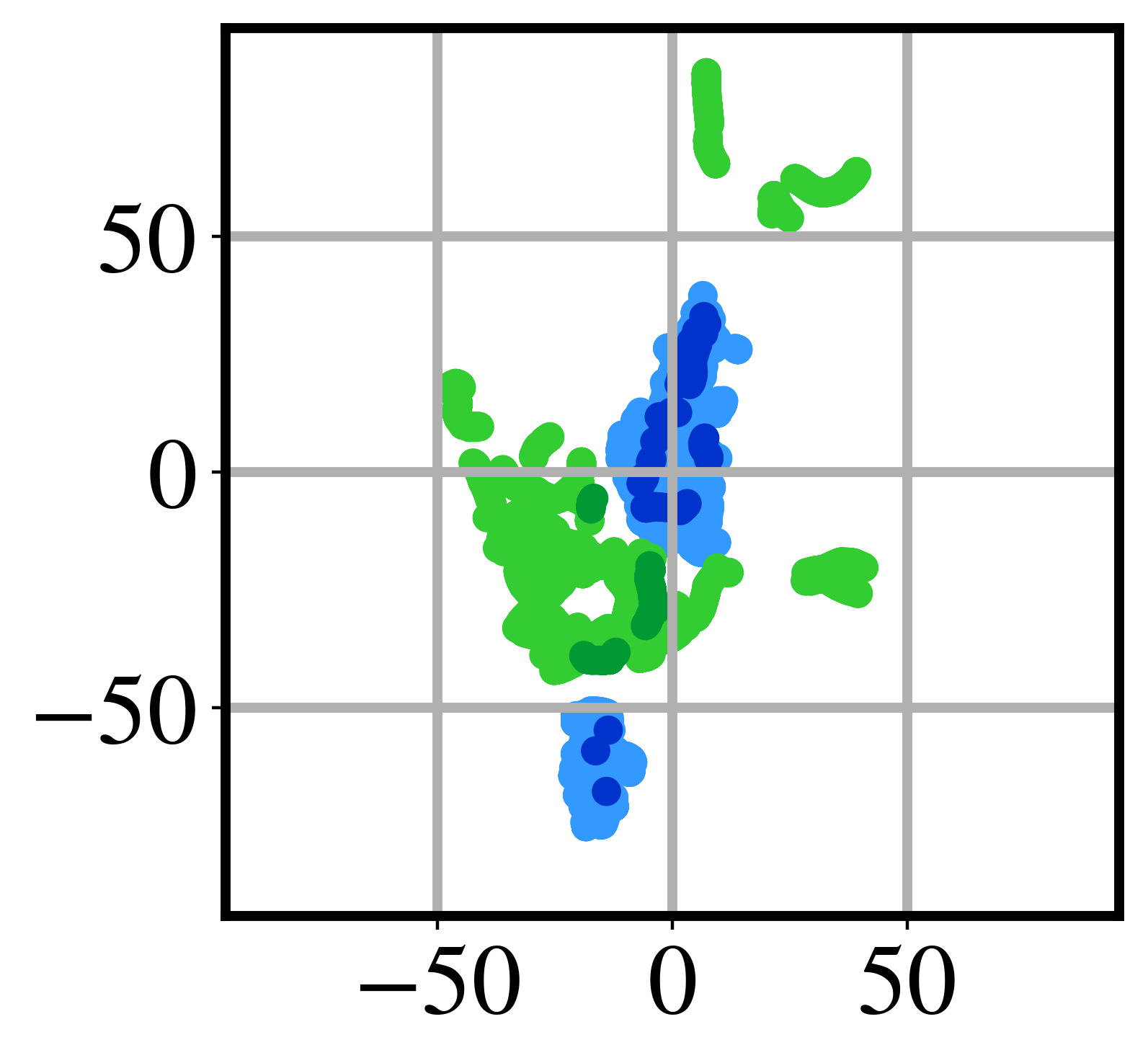}};
\node[inner sep=0pt] at (6,2)
    {\includegraphics[width=.165\textwidth]{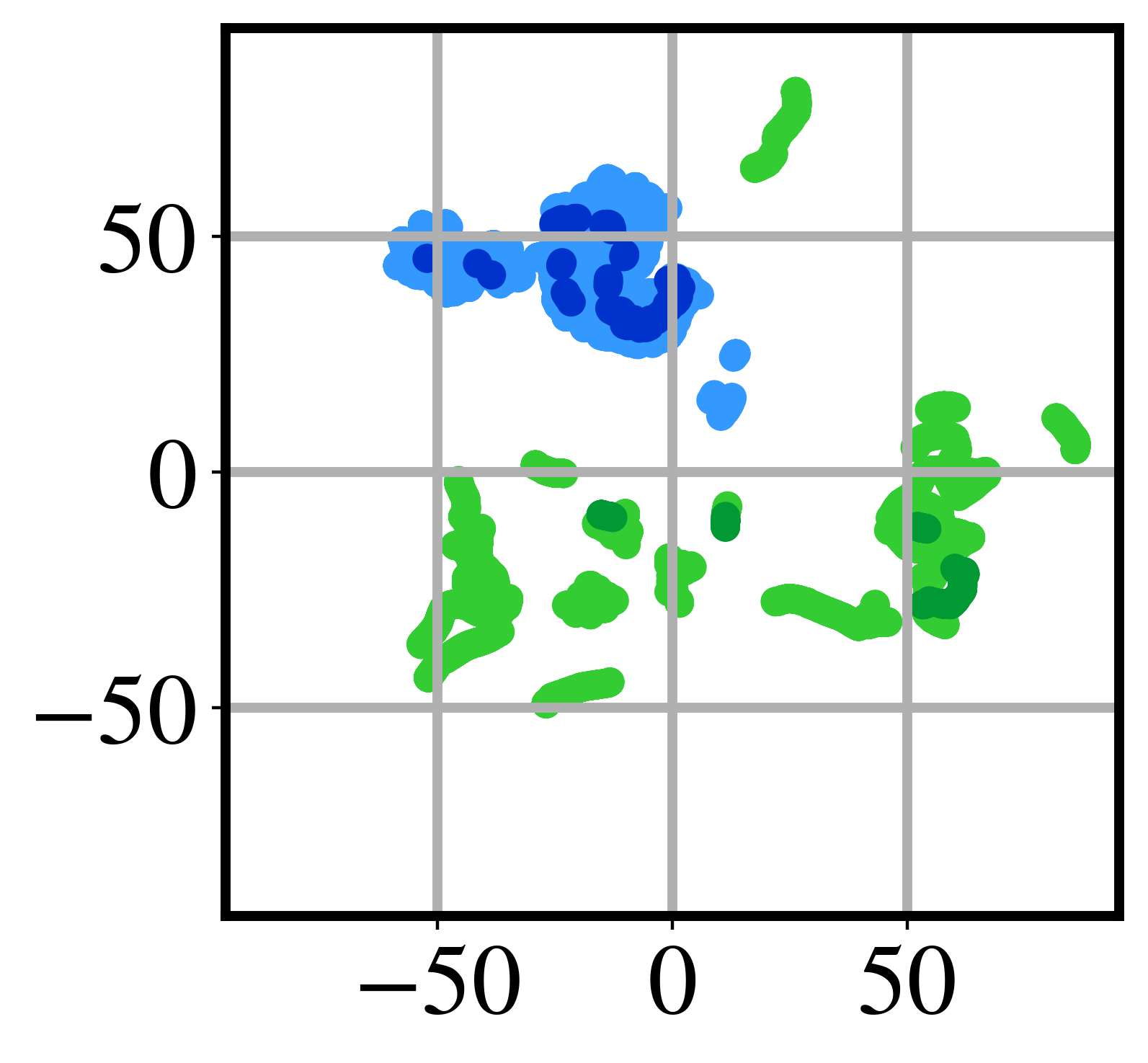}};
\node[inner sep=0pt] at (8,2)
    {\includegraphics[width=.165\textwidth]{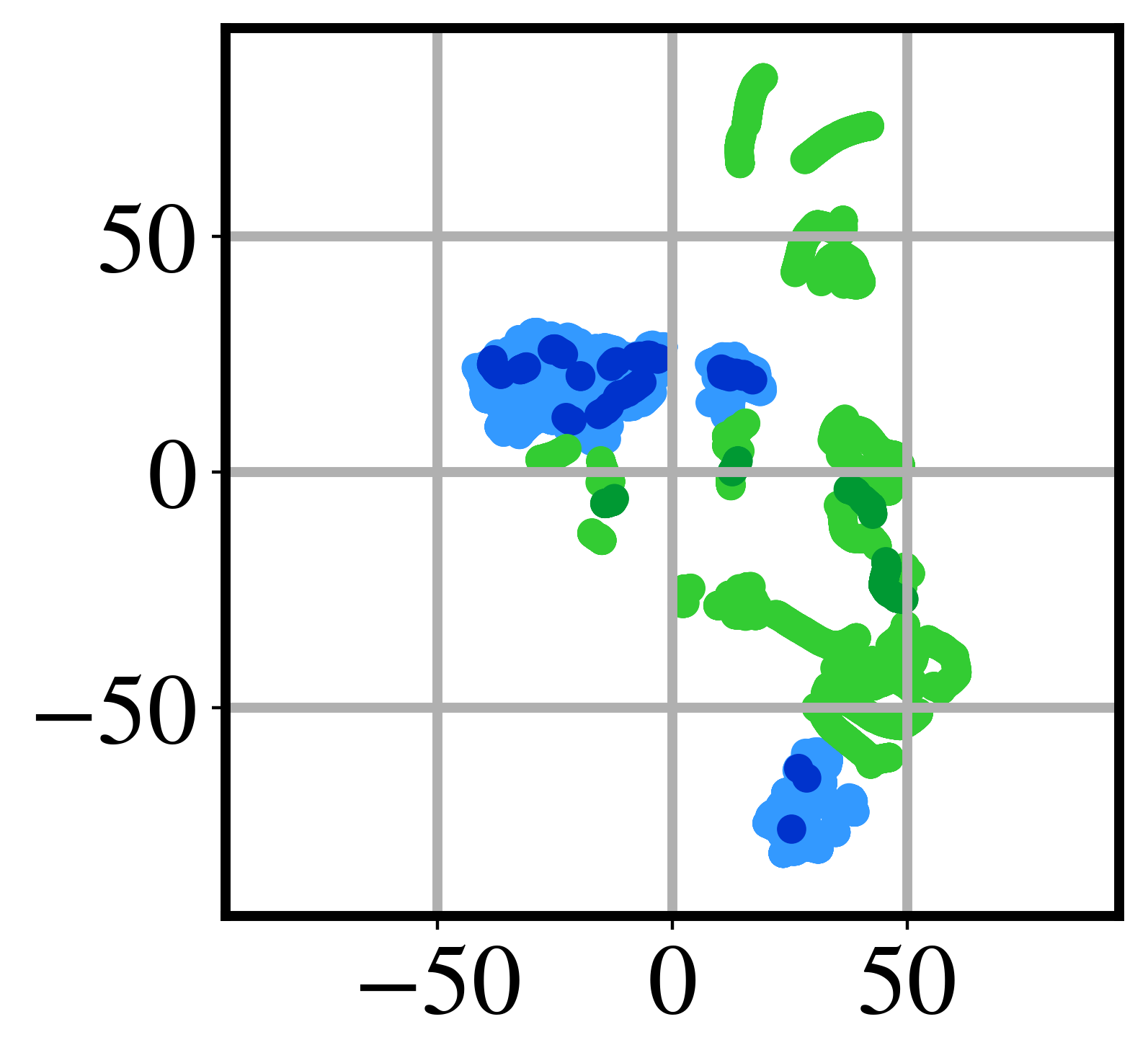}};

%% Legend
\node at (.175,3.05) {\scalebox{.85}{Epoch 10}};
\node at (2.175,3.05) {\scalebox{.85}{Epoch 15}};
\node at (4.175,3.05) {\scalebox{.85}{Epoch 20}};
\node at (6.175,3.05) {\scalebox{.85}{Epoch 25}};
\node at (8.175,3.05) {\scalebox{.85}{Epoch 30}};

\node[rotate=90] at (-1.1,2.075) {\scalebox{.85}{UNet DSL}};
\node[rotate=90] at (-1.4,.075) {\scalebox{.85}{UNet DSL}};
\node[rotate=90] at (-1.1,.075) {\scalebox{.85}{$+\mathcal{L}_{\text{Contrastive}}$}};

\node[inner sep=0pt] at (4.175,-1.35)
    {\includegraphics[width=.5\textwidth]{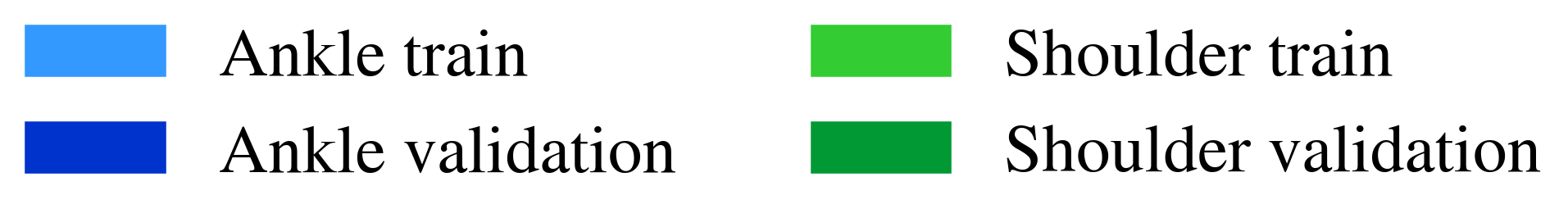}};

\draw[line width=.15mm] (1.15,-.95) rectangle (7.2,-1.7);

\end{tikzpicture}
\end{adjustbox}
  \caption{Visualization of the contrastive regularization which promotes intra-domain cohesion and inter-domain margins in embedded space during optimization. This visualization was obtained using the t-SNE algorithm in which each colored dot represented a 2D MR slice from the training or validation set of the ankle or shoulder datasets.}
  \label{fig:visualization_contrastive_regularization_supp}
\end{figure}

\end{document}